\documentclass{article}

\usepackage[final,nonatbib]{neurips_2021}

\usepackage[utf8]{inputenc} % allow utf-8 input
\usepackage[T1]{fontenc}    % use 8-bit T1 fonts
\usepackage{hyperref}       % hyperlinks
\usepackage{url}            % simple URL typesetting
\usepackage{amsfonts}       % blackboard math symbols
\usepackage{nicefrac}       % compact symbols for 1/2, etc.
\usepackage{microtype}      % microtypography
\usepackage{wrapfig}
\usepackage{graphicx}
\graphicspath{{figure}, {images}, {example}}
\DeclareGraphicsExtensions{.png,.PNG,.jpeg,.JPEG,.jpg,.JPG,.bmp,.BMP}
\usepackage{epsfig} % for figures
\usepackage{subcaption}
\usepackage[export]{adjustbox}
\usepackage{float}
\usepackage[justification=raggedright]{caption}	% makes captions ragged right - thanks to Bryce Lobdell
\usepackage{lscape} % Useful for wide tables or figures.
\usepackage{overpic}

\usepackage[lined,ruled,linesnumbered]{algorithm2e}

\SetAlFnt{\small}
\SetAlCapFnt{\small}
\SetAlCapNameFnt{\small}
\SetAlCapHSkip{0pt}
\IncMargin{-\parindent}

\usepackage{booktabs}  % Publication quality tables
\usepackage{multirow}
\usepackage{paralist}
\usepackage{enumitem}

\usepackage{mathtools}
\usepackage{amssymb,amsmath}   % Short math guide for LaTeX ftp://ftp.ams.org/pub/tex/doc/amsmath/short-math-guide.pdf

\usepackage{units}
\usepackage{xcolor}

\usepackage{comment}
\newlength\paramargin
\newlength\figmargin
\newlength\secmargin
\newlength\figcapmargin

\newcommand{\mpage}[2]
{
\begin{minipage}{#1\linewidth}\centering
#2
\end{minipage}
}

\newcommand{\heading}[1]
{
\vspace{1mm}
\noindent \textbf{#1}
}   

\newcommand{\secref}[1]{Section~\ref{sec:#1}}
\newcommand{\figref}[1]{Figure~\ref{fig:#1}} 
\newcommand{\tblref}[1]{Table~\ref{tbl:#1}}

\long\def\ignorethis#1{}

\newtheorem{theorem}{Theorem}

\usepackage{threeparttable}

\title{NeuS: Learning Neural Implicit Surfaces \\ by Volume Rendering for Multi-view Reconstruction}

\def \hku{$^\dagger$}
\def \tamu{$^\diamond$}
\def \mpi{$^\ddag$}
\def \cor{$^*$}
\author{%
  Peng Wang\hku,
  Lingjie Liu\mpi\thanks{Corresponding authors.}\hspace{4pt},
  Yuan Liu\hku,
  Christian Theobalt\mpi,
  Taku Komura\hku,
  Wenping Wang\tamu\cor
  \\
  \hku The University of Hong Kong \ \mpi Max Planck Institute for Informatics\\
  \tamu Texas A\&M University \\
  \hku \texttt{\{pwang3,yliu,taku\}@cs.hku.hk} \ \ \mpi \texttt{\{lliu,theobalt\}@mpi-inf.mpg.de} \\
  \tamu \texttt{wenping@tamu.edu}\\
}

\begin{document}

\maketitle

\begin{abstract}
We present a novel neural surface reconstruction method, called {\it NeuS}, for reconstructing objects and scenes with high fidelity from 2D image inputs. Existing neural surface reconstruction approaches, such as DVR [Niemeyer et al., 2020] and IDR [Yariv et al., 2020], require foreground mask as supervision, easily get trapped in local minima, and therefore struggle with the reconstruction of objects with severe self-occlusion or thin structures. Meanwhile, recent neural methods for novel view synthesis, such as NeRF [Mildenhall et al., 2020] and its variants, use volume rendering to produce a neural scene representation with robustness of optimization, even for highly complex objects. However, extracting high-quality surfaces from this learned implicit representation is difficult because there are not sufficient surface constraints in the representation. In NeuS, we propose to represent a surface as the zero-level set of a {\it signed distance function} (SDF) and develop a new volume rendering method to train a neural SDF representation. We observe that the conventional volume rendering method causes inherent geometric errors (i.e. bias) for surface reconstruction, and therefore propose a new formulation that is free of bias in the first order of approximation, thus leading to more accurate surface reconstruction even without the mask supervision. Experiments on the DTU dataset and the BlendedMVS dataset show that NeuS outperforms the state-of-the-arts in high-quality surface reconstruction, especially for objects and scenes with complex structures and self-occlusion.

\end{abstract}

\section{Introduction}
\label{sec:intro}
% Implict representation 

\begin{figure}[htb]
  \includegraphics[width=\linewidth]{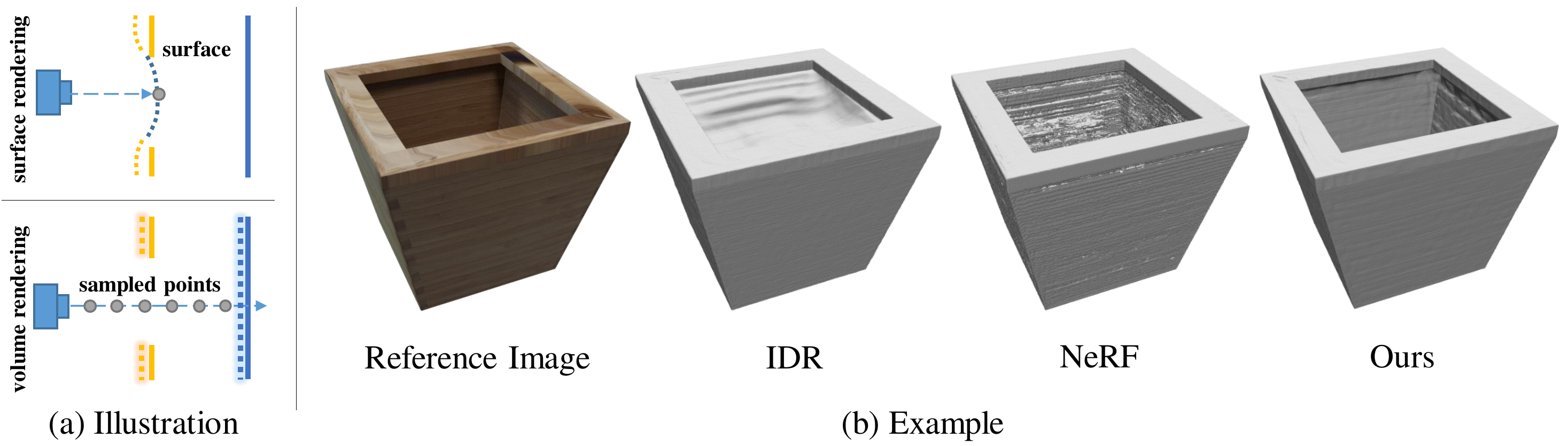}
  \caption{(a) Illustration of the surface rendering and volume rendering. (b) A toy example of bamboo planter, where there are occlusions on the top of the planter. Compared to the state-of-the-art methods, our approach can handle the occlusions and achieve better reconstruction quality. }
  \label{fig:intro_illustration}
\end{figure}

Reconstructing surfaces from multi-view images is a fundamental problem in computer vision and computer graphics. 
3D reconstruction with neural implicit representations has recently become a highly promising alternative to classical reconstruction approaches~\cite{schonberger2016pixelwise, furukawa2009accurate, barnes2009patchmatch} due to its high reconstruction quality and its potential to reconstruct complex objects that are difficult for classical approaches, such as non-Lambertian surfaces and thin structures.
Recent works represent surfaces as signed distance functions (SDF)~\cite{yariv2020multiview, zhang2021physg, kellnhofer2021neural, liu2020dist} or occupancy~\cite{niemeyer2020differentiable, oechsle2021unisurf}. 
To train their neural models, these methods use a differentiable surface rendering method to render a 3D object into images and compare them against input images for supervision.
For example, IDR~\cite{yariv2020multiview} produces impressive reconstruction results, but it fails to reconstruct objects with complex structures that causes abrupt depth changes.
The cause of this limitation is that the surface rendering method used in IDR only considers a single surface intersection point for each ray. Consequently, the gradient only exists at this single point, which is too local for effective back propagation and would get optimization stuck in a poor local minimum when there are abrupt changes of depth on images. Furthermore, object masks are needed as supervision for converging to a valid surface.
As illustrated in Fig.~\ref{fig:intro_illustration} (a) top, with the radical depth change caused by the hole, the neural network would incorrectly predict the points near the front surface to be blue, failing to find the far-back blue surface. The actual test example in Fig.~\ref{fig:intro_illustration} (b) shows that IDR fails to correctly reconstruct the surfaces near the edges with abrupt depth changes.

Recently, NeRF~\cite{mildenhall2020nerf} and its variants have explored to use a volume rendering method to learn a volumetric radiance field for novel view synthesis. 
This volume rendering approach samples multiple points along each ray and perform $\alpha$-composition of the colors of the sampled points to produce the output pixel colors for training purposes. The advantage of the volume rendering approach is that it can handle abrupt depth changes, because it considers multiple points along the ray and so all the sample points, either near the surface or on the far surface, produce gradient signals for back propagation. For example, referring Fig.~\ref{fig:intro_illustration} (a) bottom, when the near surface (yellow) is found to have inconsistent colors with the input image, the volume rendering approach is capable of training the network to find the far-back surface to produce the correct scene representation. 
However, since it is intended for novel view synthesis rather than surface reconstruction, NeRF only learns a volume density field, from which it is difficult to extract a high-quality surface.
Fig.~\ref{fig:intro_illustration} (b) shows a surface extracted as a level-set surface of the density field learned by NeRF.
Although the surface correctly accounts for abrupt depth changes, it contains conspicuous noise in some planar regions.

In this work, we present a new neural rendering scheme, called \textit{NeuS}, for multi-view surface reconstruction. NeuS uses the {\it signed distance function} (SDF) for surface representation and uses a novel volume rendering scheme to learn a neural SDF representation. Specifically, by introducing a density distribution induced by SDF, we make it possible to apply the volume rendering approach to learning an implicit SDF representation and thus have the best of both worlds, i.e. an accurate surface representation using a neural SDF model and robust network training in the presence of abrupt depth changes as enabled by volume rendering. 
Note that simply applying a standard volume rendering method to the density associated with SDF would lead to discernible bias (i.e. inherent geometric errors) in the reconstructed surfaces. This is a new and important observation that we will elaborate later. Therefore we propose a novel volume rendering scheme to ensure unbiased surface reconstruction in the first-order approximation of SDF.
Experiments on both DTU dataset and BlendedMVS dataset demonstrated that NeuS is capable of reconstructing complex 3D objects and scenes with severe occlusions and delicate structures, even without foreground masks as supervision. It outperforms the state-of-the-art neural scene representation methods, namely IDR~\cite{yariv2020multiview} and NeRF~\cite{mildenhall2020nerf}, in terms of reconstruction quality.

\section{Related Works}

\heading{Classical Multi-view Surface and Volumetric Reconstruction.}
Traditional multi-view 3D reconstruction methods can be roughly classified into two categories:  point- and surface-based reconstruction  \cite{barnes2009patchmatch,furukawa2009accurate,galliani2016gipuma, schonberger2016pixelwise} and volumetric reconstruction \cite{de1999poxels,broadhurst2001probabilistic, seitz1999photorealistic}. 
Point- and surface-based reconstruction methods estimate the depth map of each pixel by exploiting inter-image photometric consistency~\cite{furukawa2009accurate} and then fuse the depth maps into a global dense point cloud~\cite{Merrell2007,Zach2007}. The surface reconstruction is usually done as a post processing with methods like screened Poisson surface reconstruction~\cite{poisson}.
The reconstruction quality heavily relies on the quality of correspondence matching, and the difficulties in matching correspondence for objects without rich textures often lead to severe artifacts and missing parts in the reconstruction results.
Alternatively, volumetric reconstruction methods circumvent the difficulty of explicit correspondence matching by estimating occupancy and color in a voxel grid from multi-view images and evaluating the color consistency of each voxel. Due to limited achievable voxel resolution, these methods cannot achieve high accuracy.

\heading{Neural Implicit Representation.}
Some methods enforce 3D understanding in a deep learning framework by introducing inductive biases. These inductive biases can be explicit representations, such as voxel grids~\cite{kar2017learning, choy20163d, xie2019pix2vox}, point cloud~\cite{fan2017point, mandikal20183dlmnet, lin2018learning}, meshes~\cite{wang2018pixel2mesh, wen2019pixel2mesh++, kato2018neural}, and implicit representations. The implicit representations encoded by a neural network has gained a lot of attention recently, since it is continuous and can achieve high spatial resolution.
This representation has been applied successfully to shape representation~\cite{mescheder2019occupancy,Michalkiewicz_2019_ICCV,park2019deepsdf,8953765,atzmon2020sal,gropp2020implicit,yifan2020iso,Peng2020ConvolutionalON}, novel view synthesis~\cite{sitzmann2019scene,lombardi2019neural,kaza2019differentiable,mildenhall2020nerf,liu2020neural,saito2019pifu,saito2020pifuhd, trevithick2020grf, sitzmann2019deepvoxels} and multi-view 3D reconstruction \cite{yariv2020multiview,niemeyer2020differentiable,kellnhofer2021neural,jiang2020sdfdiff, liu2020dist}.

Our work mainly focuses on learning implicit neural representation encoding both geometry and appearance in 3D space from 2D images via classical rendering techniques. 
Limited in this scope, the related works can be roughly categorized based on the rendering techniques used, i.e. surface rendering based methods and volume rendering based methods. Surface rendering based methods~\cite{niemeyer2020differentiable, kellnhofer2021neural, yariv2020multiview, liu2020dist} assume that the color of ray only relies on the color of an intersection of the ray with the scene geometry, which makes the gradient only backpropagated to a local region near the intersection. 
Therefore, such methods struggle with reconstructing complex objects with severe self-occlusions and sudden depth changes. Furthermore, they usually require object masks as supervision. 
On the contrary, our method performs well for such challenging cases without the need of masks. 

Volume rendering based methods, such as NeRF\cite{mildenhall2020nerf}, render an image by $\alpha$-compositing colors of the sampled points along each ray. As explained in the introduction, it can handle sudden depth changes and synthesize high-quality images.
However, extracting high-fidelity surface from the learned implicit field is difficult because the density-based scene representation lacks sufficient constraints on its level sets.
In contrast, our method combines the advantages of surface rendering based and volume rendering based methods by constraining the scene space as a signed distance function but applying volume rendering to train this representation with robustness.
UNISURF~\cite{oechsle2021unisurf}, a concurrent work, also learns an implicit surface via volume rendering. It improves the reconstruction quality by shrinking the sample region of volume rendering during the optimization. Our method differs from UNISURF in that UNISURF represents the surface by occupancy values,
while our method represents the scene by an SDF and thus can naturally extract the surface as the zero-level set of it, yielding better reconstruction accuracy than UNISURF, as will be seen later in the experiment section.

\section{Method}

Given a set of posed images $\left\{\mathcal{I}_k\right\}$ of a 3D object, our goal is to reconstruct the surface $\mathcal{S}$ of it.
The surface is represented by the zero-level set of a neural implicit SDF. In order to learn the weights of the neural network, we developed a novel volume rendering method to render images from the implicit SDF and minimize the difference between the rendered images and the input images. This volume rendering approach ensures robust optimization in NeuS for reconstructing objects of complex structures.

\subsection{Rendering Procedure}
\label{sec:architecture}
\heading{Scene representation.} With NeuS, the scene of an object to be reconstructed is represented by two functions: $f:\mathbb{R}^3\to \mathbb{R}$ that maps a spatial position ${\bf x}\in \mathbb{R}^3$ to its signed distance to the object, and $c:\mathbb{R}^3 \times \mathbb{S}^2 \to \mathbb{R}^3$ that encodes the color associated with a point ${\bf x} \in \mathbb{R}^3$ and a viewing direction ${\bf v} \in \mathbb{S}^2$. Both functions are encoded by Multi-layer Perceptrons~(MLP). The surface $\mathcal{S}$ of the object is represented by the zero-level set of its SDF, that is,
\begin{equation}
    \mathcal{S} = \left\{ \mathbf{x} \in \mathbb{R}^3 | f(\mathbf{x}) = 0 \right\}.
\end{equation}

In order to apply a volume rendering method to training the SDF network, we first introduce a probability density function $\phi_s(f({\bf x}))$, called {\em S-density}, where $f({\bf x})$, ${\bf x} \in \mathbb{R}^3$, is the signed distance function and $\phi_s(x) = se^{-sx}/(1 + e^{-sx})^2$, commonly known as the {\em logistic density distribution}, is the derivative of the Sigmoid function $\Phi_s(x) = (1 + e^{-sx})^{-1}$, i.e., $\phi_s(x) = \Phi_s'(x)$. In principle $\phi_s(x)$ can be any unimodal (i.e. bell-shaped) density distribution centered at $0$; here we choose the logistic density distribution for its computational convenience. Note that the standard deviation of $\phi_s(x)$ is given by $1/s$, which is also a trainable parameter, that is, $1/s$ approaches to zero as the network training converges.

Intuitively, the main idea of NeuS is that, with the aid of the S-density field $\phi_s(f({\bf x}))$, volume rendering is used to train the SDF network with only 2D input images as supervision. Upon successful minimization of a loss function based on this supervision, the zero-level set of the network-encoded SDF is expected to represent an accurately reconstructed surface $\mathcal{S}$, with its induced S-density $\phi_s(f({\bf x}))$ assuming prominently high values near the surface.

\heading{Rendering.} To learn the parameters of the neural SDF and color field, we advise a volume rendering scheme to render images from the proposed SDF representation.
Given a pixel, we denote the ray emitted from this pixel as $\{\mathbf{p}(t) = \mathbf{o} + t\mathbf{v} | t\geq 0\}$, where $\mathbf{o}$ is the center of the camera and $\mathbf{v}$ is the unit direction vector of the ray. We accumulate the colors along the ray by
\begin{equation}
    C(\mathbf{o}, \mathbf{v}) = \int_{0}^{+\infty}w(t)c(\mathbf{p}(t), \mathbf{v}){\rm d}t,
    \label{eq:int}
\end{equation}
where $C(\mathbf{o}, \mathbf{v})$ is the output color for this pixel, $w(t)$ a weight for the point $\mathbf{p}(t)$, and $c(\mathbf{p}(t),\mathbf{v})$ the color at the point $\mathbf{p}$ along the viewing direction $\mathbf{v}$.

{\bf Requirements on weight function}.
The key to learn an accurate SDF representation from 2D images is to build an appropriate connection between output colors and SDF, i.e., to derive an appropriate weight function $w(t)$ on the ray based on the SDF $f$ of the scene. In the following, we list the requirements on the weight function $w(t)$.

\begin{enumerate}
    \item \textbf{Unbiased}. Given a camera ray $\mathbf{p}(t)$, $w(t)$ attains a locally maximal value at a surface intersection point $\mathbf{p}(t^*)$, i.e. with $f(\mathbf{p}(t^*))=0$, that is, the point $\mathbf{p}(t^*)$ is on the zero-level set of the SDF $\mathbf({\bf x})$.
    \item \textbf{Occlusion-aware}. Given any two depth values $t_0$ and $t_1$ satisfying $f(t_0)=f(t_1)$, $w(t_0) > 0$, $w(t_1) > 0$, and $t_0<t_1$, there is $w(t_0)>w(t_1)$. That is, when two points have the same SDF value (thus the same SDF-induced S-density value), the point nearer to the view point should have a larger contribution to the final output color than does the other point. 
\end{enumerate}

\begin{wrapfigure}{R}{0.6\textwidth}
\centering{
  \vspace{0pt}
  \includegraphics[width=\linewidth]{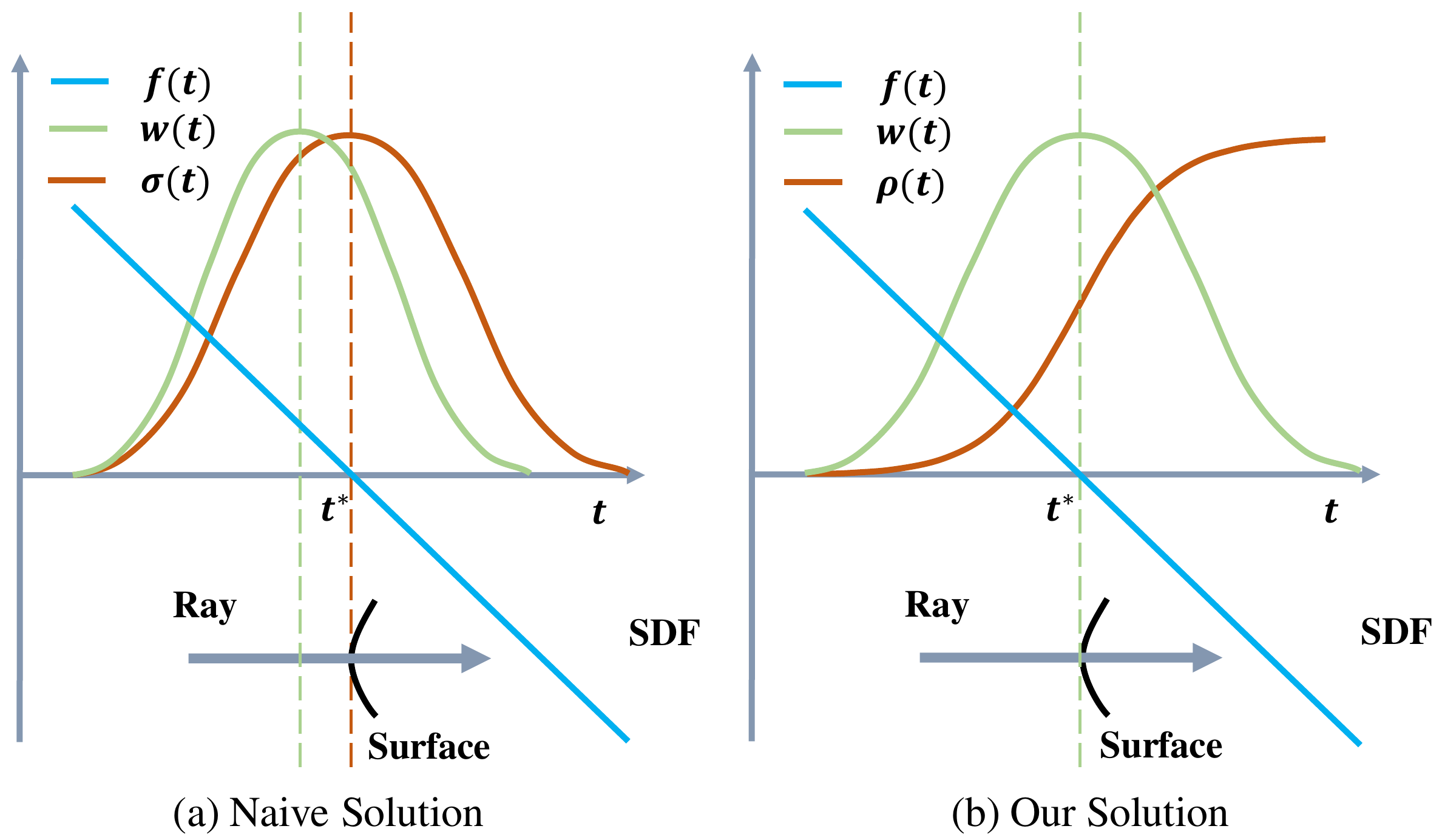}
  \caption{Illustration of (a) weight bias of naive solution, and (b) the weight function defined in our solution, which is unbiased in the first-order approximation of SDF.}
  \vspace{-10pt}
  \label{fig:method_bias}
  }
\end{wrapfigure}

An unbiased weight function $w(t)$ guarantees that the intersection of the camera ray with the zero-level set of SDF contributes most to the pixel color.
The occlusion-aware property ensures that when a ray sequentially passes multiple surfaces, the rendering procedure will correctly use the color of the surface nearest to the camera to compute the output color.

Next, we will first introduce a naive way of defining the weight function $w(t)$ that directly using the standard pipeline of volume rendering, and explain why it is not appropriate for reconstruction before introducing our novel construction of $w(t)$.

\textbf{Naive solution}. To make the weight function occlusion-aware, a natural solution is based on the standard volume rendering formulation~\cite{mildenhall2020nerf} which defines the weight function by 
\begin{equation}
w(t)=T(t)\sigma(t),
\label{eq:volume}
\end{equation}
where $\sigma(t)$ is the so-called {\it volume density} in classical volume rendering and $T(t)=\exp(-\int_{0}^{t}\sigma(u){\rm d}u)$ here denotes the {\it accumulated transmittance} along the ray. To adopt the standard volume density formulation~\cite{mildenhall2020nerf}, here $\sigma(t)$ is set to be equal to the S-density value, i.e. $\sigma(t) = \phi_s(f(\mathbf{p}(t)))$ and the weight function $w(t)$ is computed by Eqn.~\ref{eq:volume}. Although the resulting weight function is occlusion-aware, it is biased as it introduces inherent errors in the reconstructed surfaces. As illustrated in Fig.~\ref{fig:method_bias} (a), the weight function $w(t)$ attains a local maximum at a point before the ray reaches the surface point $\mathbf{p}(t^*)$, satisfying $f(\mathbf{p}(t^*))=0$. This fact will be proved in the supplementary material.

\textbf{Our solution}. To introduce our solution, we first introduce a straightforward way to construct an unbiased weight function, which directly uses the normalized S-density as weights

\begin{equation}
    w(t)=\frac{\phi_s (f(\mathbf{p}(t)))}{\int_{0}^{+\infty} \phi_s(f(\mathbf{p}(u))){\rm d}u}.
    \label{eq:trivial1}
\end{equation}

This construction of weight function is unbiased, but not occlusion-aware. For example, if the ray penetrates two surfaces, the SDF function $f$ will have two zero points on the ray, which leads to two peaks on the weight function $w(t)$ and the resulting weight function will equally blend the colors of two surfaces without considering occlusions.

To this end, now we shall design the weight function $w(t)$ that is both occlusion-aware and unbiased in the first-order approximation of SDF, based on the aforementioned straightforward construction. To ensure an occlusion-aware property of the weight function $w(t)$, we will still follow the basic framework of volume rendering as Eqn.~\ref{eq:volume}. However, different from the conventional treatment as in naive solution above, we define our function $w(t)$ from the S-density in a new manner. We first define an opaque density function $\rho(t)$, which is the counterpart of the volume density $\sigma$ in standard volume rendering. Then we compute the new weight function $w(t)$ by 

\begin{equation}
w(t) = T(t)\rho(t), \;\; {\rm where } \  T(t) =  \exp\left(-\int_{0}^{t}\rho(u){\rm d}u\right).
\label{eq:new_weight}
\end{equation}

\textbf{How we derive opaque density $\rho$}.
% We first consider a simple case where there is only one surface intersection, and the surface is simply a plane.
We first consider a simple ideal case where there is only one surface intersection, and the surface is simply a plane that approaches infinitely far off the camera. Since Eqn.~\ref{eq:trivial1} indeed satisfies the above requirements under this assumption, we derive the underlying opaque density $\rho$ corresponding to the weight definition of Eqn.~\ref{eq:trivial1} using the framework of volume rendering. Then we will generalize this opaque density to the general case of multiple surface intersections.

Specifically, in the simple case of a single plane intersection, it is easy to see that the signed distance function $f(\mathbf{p}(t))$ is $-|\cos(\theta)|\cdot(t - t^*)$, where $f(\mathbf{p}(t^*)) = 0$, and $\theta$ is the angle between the view direction $\mathbf{v}$ and the outward surface normal vector $\mathbf{n}$. Because the surface is assumed a plane, $|\cos(\theta)|$ is a constant. It follows from Eqn.~\ref{eq:trivial1} that
\begin{equation}
\begin{split}
w(t) =& {\lim_{t^* \to +\infty}\frac{\phi_s(f(\mathbf{p}(t)))}{\int_{0}^{+\infty}\phi_s(f(\mathbf{p}(u))){\rm d} u}}\\
=& {\lim_{t^* \to +\infty}\frac{\phi_s(f(\mathbf{p}(t)))}{\int_{0}^{+\infty}\phi_s(-|\cos(\theta)|(u - t^*)){\rm d} u}}\\
% =& {\lim_{t^* \to +\infty}}\frac{\phi_s(f(\mathbf{p}(t)))}{|\cos(\theta)|^{-1}\cdot\blue{\int_{0}^{+\infty}}\phi_s(u - t^*){\rm d} u}\\
=& {\lim_{t^* \to +\infty}\frac{\phi_s(f(\mathbf{p}(t)))}{\int_{-t^*}^{+\infty}\phi_s(-|\cos(\theta)| u^*){\rm d} u^*}}\\
=& {\lim_{t^* \to +\infty}\frac{\phi_s(f(\mathbf{p}(t)))}{|\cos(\theta)|^{-1}\int_{-|\cos(\theta)|t^*}^{+\infty} \phi_s(\hat{u}) {\rm d} \hat{u}}} \\
=& |\cos(\theta)|\phi_s(f(\mathbf{p}(t))).
\end{split}
\end{equation}
Recall that the weight function within the framework of volume rendering is given by $w(t) = T(t)\rho(t)$, where $T(t) = \exp(-\int_{0}^{t}\rho(u){\rm d}u)$ denotes the {\it accumulated transmittance}. Therefore, to derive $\rho(t)$, we have
\begin{equation}
    T(t)\rho(t) = |\cos(\theta)|\phi_s(f(\mathbf{p}(t))).
\end{equation}
Since $T(t) = \exp(-\int_{0}^{t}\rho(u){\rm d}u)$, it is easy to verify that $T(t)\rho(t) = -\frac{{\rm d} T}{{\rm d} t}(t)$. Further, note that $|\cos(\theta)|\phi_s(f(\mathbf{p}(t))) = -\frac{{\rm d} \Phi_s}{{\rm d} t}(f(\mathbf{p}(t)))$. It follows that $\frac{{\rm d} T}{{\rm d} t}(t) =  \frac{{\rm d} \Phi_s}{{\rm d} t}(f(\mathbf{p}(t)))$. Integrating both sides of this equation yields
\begin{equation}
    T(t) = \Phi_s(f(\mathbf{p}(t))).
\end{equation}
Taking the logarithm and then differentiating both sides, we have
\begin{equation}
\begin{split}
{\int_{0}^{t}}\rho(u){\rm d}u =& -\ln(\Phi_s(f(\mathbf{p}(t)))) \\
\Rightarrow \rho(t) =& \frac{-\frac{{\rm d} \Phi_s}{{\rm d} t}(f(\mathbf{p}(t)))}{\Phi_s(f(\mathbf{p}(t)))}.
\end{split}
\label{eq:intuition}
\end{equation}

\begin{wrapfigure}{R}{0.4\textwidth}
\centering{
  \vspace{0pt}
  \includegraphics[width=\linewidth]{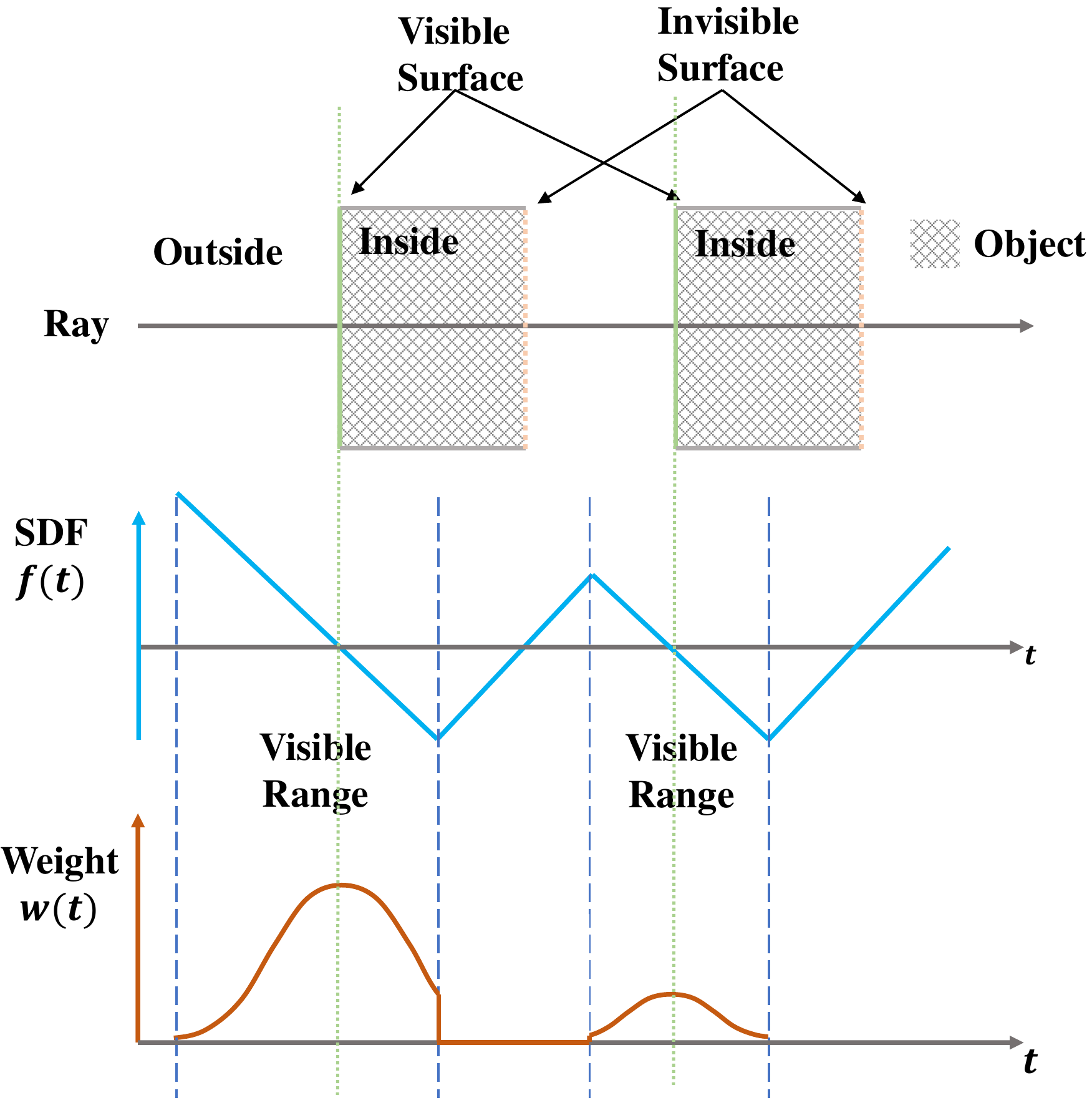}
  \caption{Illustration of weight distribution in case of multiple surface intersection.}
  \vspace{-20pt}
  \label{fig:method_illustration}
  }
\end{wrapfigure}

This is the formula of the opaque density $\rho(t)$ in the ideal case of single plane intersection. The weight function $w(t)$ induced by $\rho(t)$ is shown in \figref{method_bias}(b). Now we generalize the opaque density to the general case where there are multiple surface intersections along the ray $\mathbf{p}(t)$. In this case, $-\frac{{\rm d}\Phi_s}{{\rm d} t}(f(\mathbf{p}(t)))$ becomes negative on the segment of the ray with increasing SDF values. Thus we clip it against zero to ensure that the value of $\rho$ is always non-negative. This gives the following opaque density function $\rho(t)$ in general cases.

\begin{equation}
\begin{split}
\rho(t) =& \max\left(\frac{-\frac{{\rm d}\Phi_s}{{\rm d} t}(f(\mathbf{p}(t)))}{\Phi_s(f(\mathbf{p}(t)))}, 0\right).
\end{split}
\label{eq:sigma}
\end{equation}

Based on this equation, the weight function $w(t)$ can be computed with standard volume rendering as in Eqn. \ref{eq:new_weight}. The illustration in the case of multiple surface intersection is shown in \figref{method_illustration}.

The following theorem states that in general cases (i.e., including both single surface intersection and multiple surface intersections) the weight function defined by Eqn.~\ref{eq:sigma} and Eqn.~\ref{eq:new_weight} is unbiased in the first-order approximation of SDF. The proof is given in the supplementary material.

\begin{theorem}
    Suppose that a smooth surface $\mathbb{S}$ is defined by the zero-level set of the signed distance function $f(\mathbf{x})=0$, and a ray $\mathbf{p}(t)=\mathbf{o}+t \mathbf{v}$ enters the surface $\mathbb{S}$ from outside to inside, with the intersection point at $\mathbf{p}(t^*)$, that is, $f(\mathbf{p}(t^*))=0$ and there exists an interval $[t_l, t_r]$ such that $ t^* \in [t_l, t_r]$ and $f(\mathbf{p}(t))$ is monotonically decreasing in $[t_l, t_r]$. Suppose that in this local interval $[t_l, t_r]$, the surface can be tangentially approximated by a sufficiently small planar patch, i.e., $\nabla\mathbf{f}$ is regarded as fixed.
    Then, the weight function $w(t)$ computed by Eqn.~\ref{eq:sigma} and Eqn. \ref{eq:new_weight} in $[t_l, t_r]$ attains its maximum at $t^*$.
\end{theorem}

\textbf{Discretization}.
To obtain discrete counterparts of the opacity and weight function, we adopt the same approximation scheme as used in NeRF~\cite{mildenhall2020nerf},
This scheme samples $n$ points $\{\mathbf{p}_i=\mathbf{o}+t_i\mathbf{v}|i=1,...,n,t_i<t_{i+1}\}$ along the ray to compute the approximate pixel color of the ray as 
\begin{equation}
    \hat{C} = \sum_{i=1}^{n} T_i\alpha_i c_i,
\end{equation}
where $T_i$ is the discrete {\it accumulated transmittance} defined by $T_i=\prod_{j=1}^{i-1}(1 - \alpha_j)$, and $\alpha_i$ is discrete opacity values defined by 
\begin{equation}
\alpha_i=1-\exp\left(-\int_{t_i}^{t_{i+1}}\rho(t){\rm d}t\right),
\label{discrete_alpha_0}
\end{equation}
which can further be shown to be 
\begin{equation}
\alpha_i = \max \left (\frac{\Phi_s(f(\mathbf{p}(t_{i}))) - \Phi_s(f(\mathbf{p}(t_{i+1})))}{\Phi_s(f(\mathbf{p}(t_i)))}, 0 \right).
\label{discrete_alpha}
\end{equation}
The detailed derivation of this formula for $\alpha_i$ is given in the supplementary material.

%%%%%%%%%%%%%%%%%%%%%

\subsection{Training}
To train NeuS, we minimize the difference between the rendered colors and the ground truth colors, without any 3D supervision. 
Besides colors, we can also utilize the masks for supervision if provided.

Specifically, we optimize our neural networks and inverse standard deviation $s$ by randomly sampling a batch of pixels and their corresponding rays in world space $P = \left\{ C_k, M_k, \mathbf{o}_k, \mathbf{v}_k\right\}$, where $C_k$ is its pixel color and $M_k \in \{0, 1\}$ is its optional mask value, from an image in every iteration. We assume the point sampling size is $n$ and the batch size is $m$.
The loss function is defined as 
\begin{equation}
    \mathcal{L} = \mathcal{L}_{color} + \lambda\mathcal{L}_{reg} + \beta\mathcal{L}_{mask}.
\end{equation}
The color loss $\mathcal{L}_{color}$ is defined as
\begin{equation}
    \mathcal{L}_{color} = \frac{1}{m}\sum_{k}\mathcal{R}(\hat{C}_k, C_{k}).
\end{equation}
Same as IDR\cite{yariv2020multiview}, we empirically choose $\mathcal{R}$ as L1 loss, which in our observation is robust to outliers and stable in training.

We add an Eikonal term~\cite{gropp2020implicit} on the sampled points to regularize the SDF of $f_\theta$ by
\begin{equation}
    \mathcal{L}_{reg} = \frac{1}{nm}\sum_{k,i}(\|\nabla f(\hat{\mathbf{p}}_{k,i})\|_2 - 1)^2.
\end{equation}
The optional mask loss $\mathcal{L}_{mask}$ is defined as
\begin{equation}
    \mathcal{L}_{mask} = \text{BCE}(M_k, \hat{O}_k),
\end{equation}
where $\hat{O}_k = \sum_{i=1}^{n}T_{k,i}\alpha_{k, i}$ is the sum of weights along the camera ray, and $\text{BCE}$ is the binary cross entropy loss.

\textbf{Hierarchical sampling}.
In this work, we follow a similar hierarchical sampling strategy as in NeRF~\cite{mildenhall2020nerf}. We first uniformly sample the points on the ray and then iteratively conduct importance sampling on top of the coarse probability estimation. The difference is that, unlike NeRF which simultaneously optimizes a coarse network and a fine network, we only maintain one network, where the probability in coarse sampling is computed based on the S-density $\phi_s(f(\mathbf{x}))$ with fixed standard deviations while the probability of fine sampling is computed based on $\phi_s(f(\mathbf{x}))$ with the learned $s$. Details of hierarchical sampling strategy are provided in supplementary materials.

\begin{figure}[!b]
    \rotatebox[origin=C]{90}{\parbox{20mm}{\centering \small Scan 24 \\ (DTU)}} 
  \mpage{0.22}{\includegraphics[width=\linewidth]{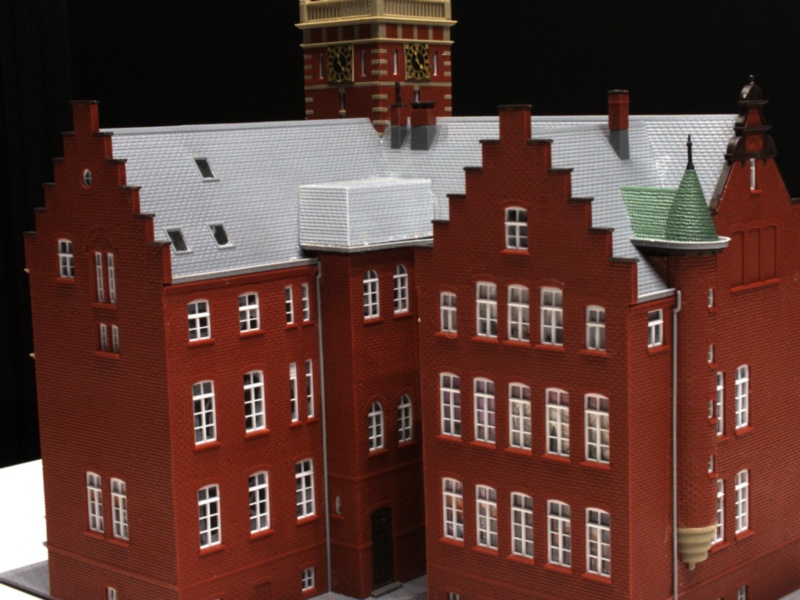}}
  \mpage{0.22}{\includegraphics[width=\linewidth]{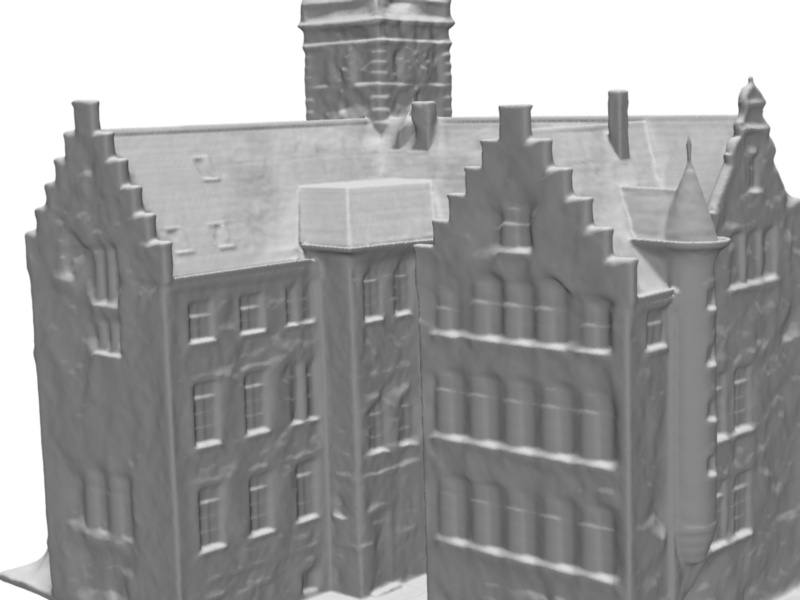}}
  \mpage{0.22}{\includegraphics[width=\linewidth]{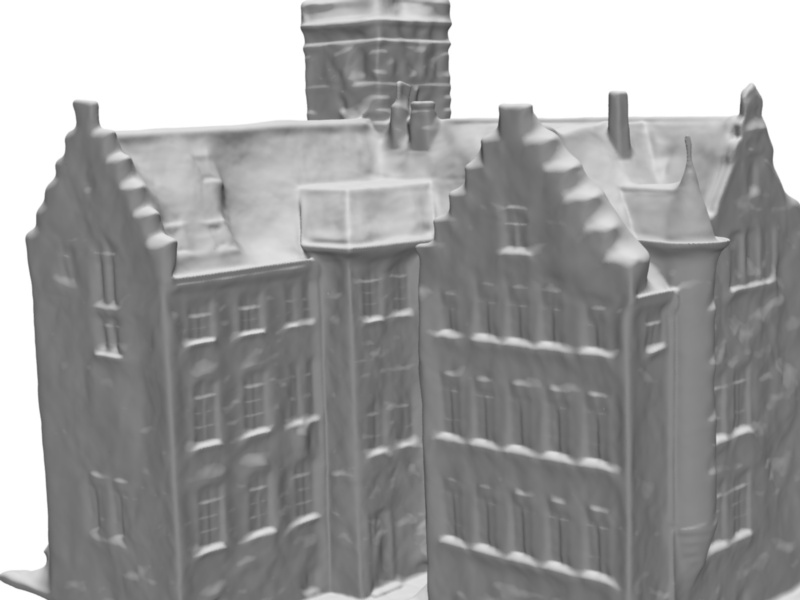}}
  \mpage{0.22}{\includegraphics[width=\linewidth]{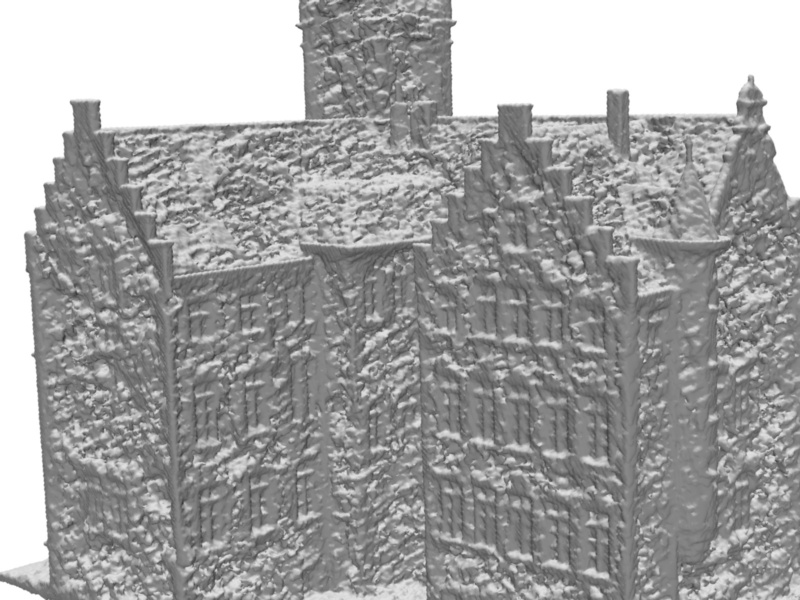}}
  \\
  \rotatebox[origin=C]{90}{\parbox{20mm}{\centering \small Scan 37 \\ (DTU)}} 
  \mpage{0.22}{\includegraphics[width=\linewidth]{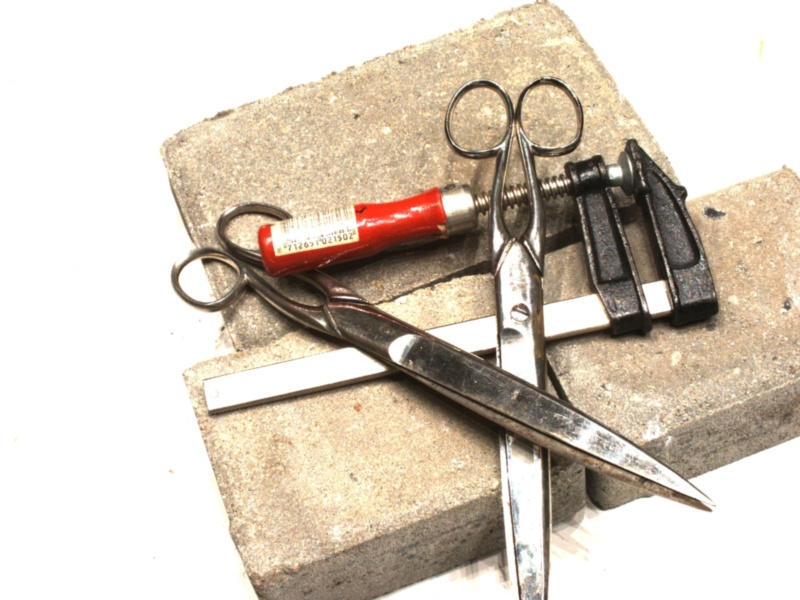}}
  \mpage{0.22}{\includegraphics[width=\linewidth]{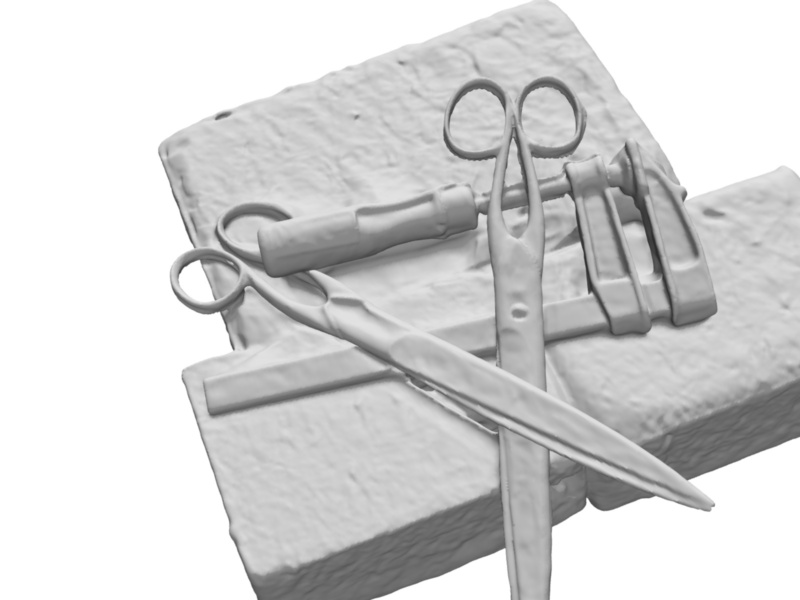}}
  \mpage{0.22}{\includegraphics[width=\linewidth]{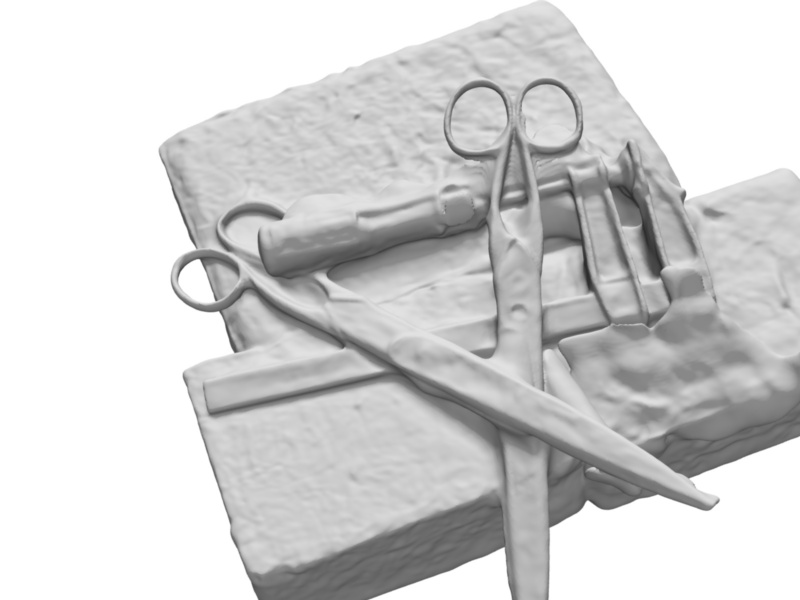}}
  \mpage{0.22}{\includegraphics[width=\linewidth]{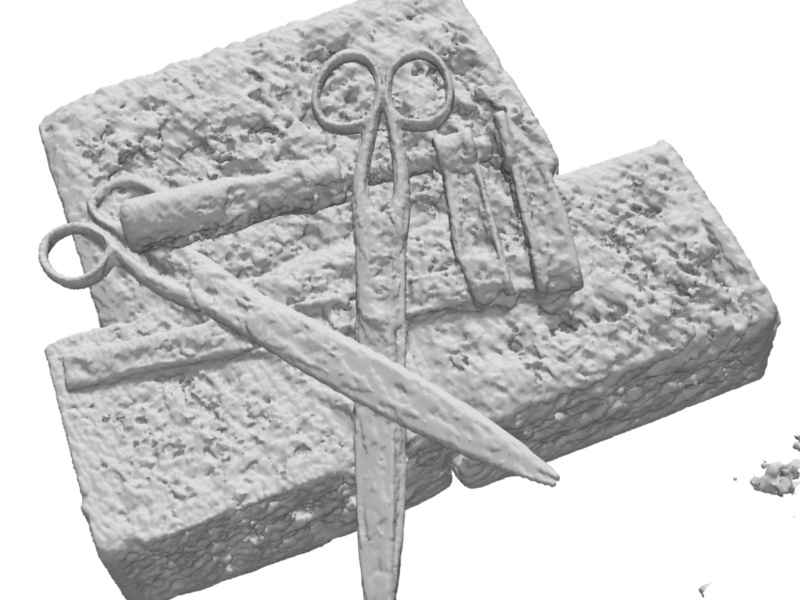}}
 \\
   \rotatebox[origin=C]{90}{\parbox{20mm}{\centering \small Dog \\ (BlendedMVS)}} 
 \mpage{0.22}{\includegraphics[width=\linewidth]{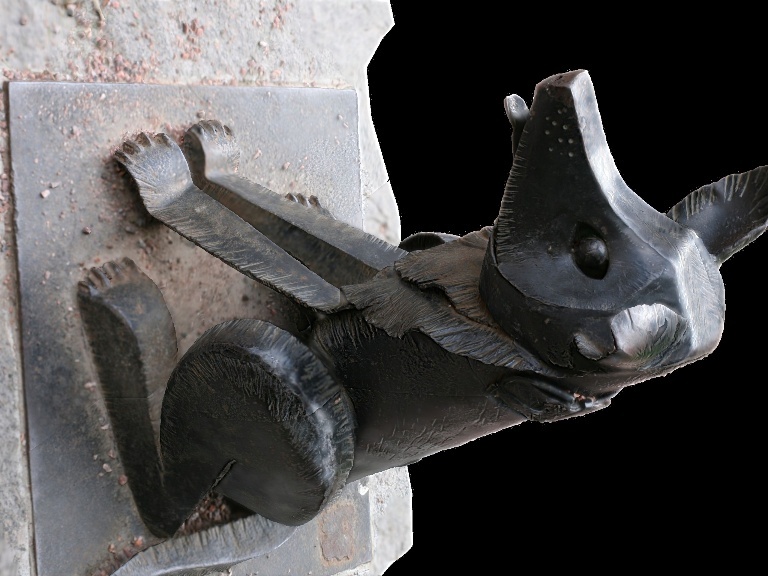}}
  \mpage{0.22}{\includegraphics[width=\linewidth]{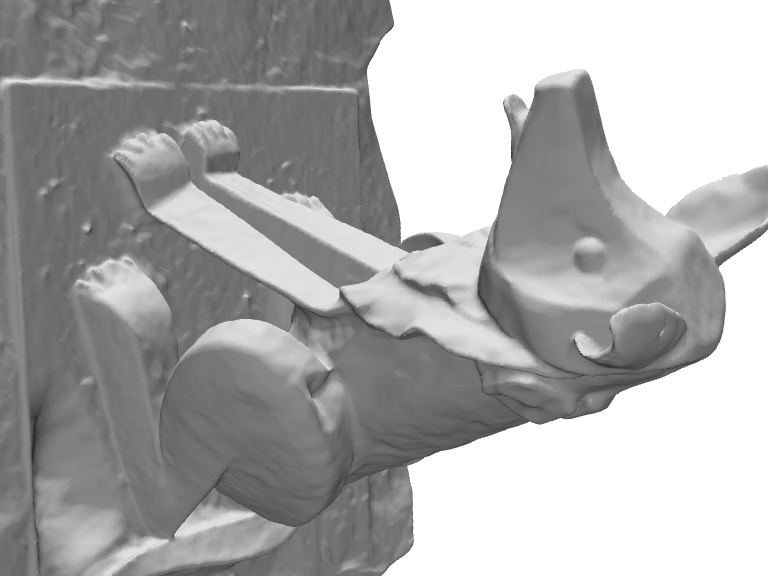}}
  \mpage{0.22}{\includegraphics[width=\linewidth]{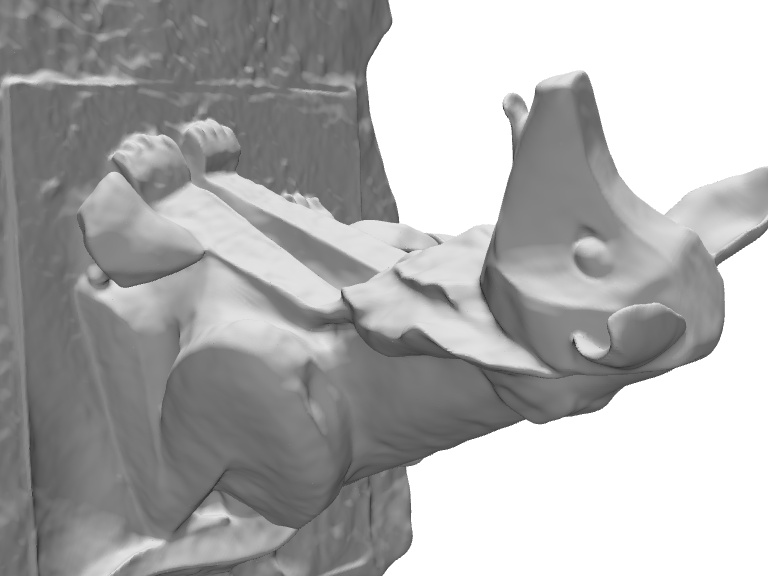}}
  \mpage{0.22}{\includegraphics[width=\linewidth]{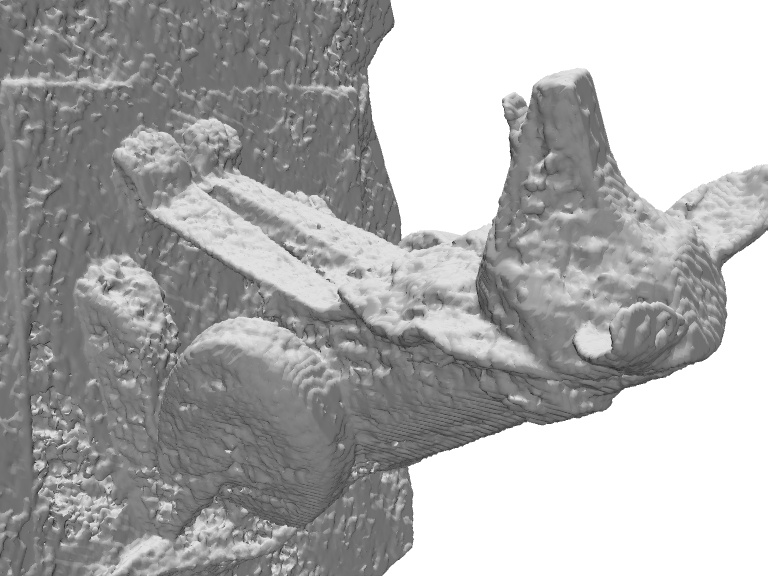}}
 \\
   \rotatebox[origin=C]{90}{\parbox{20mm}{\centering \small Stone \\ (BlendedMVS)}} 
 \mpage{0.22}{\includegraphics[width=\linewidth]{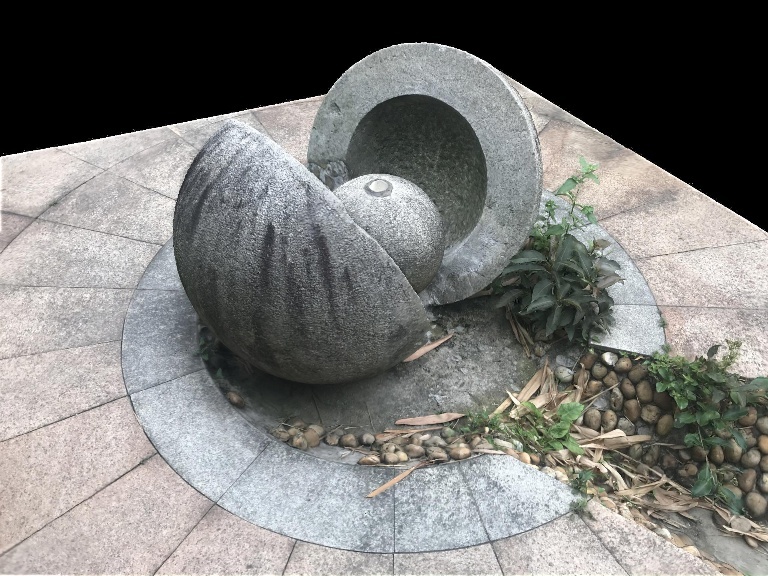}}
  \mpage{0.22}{\includegraphics[width=\linewidth]{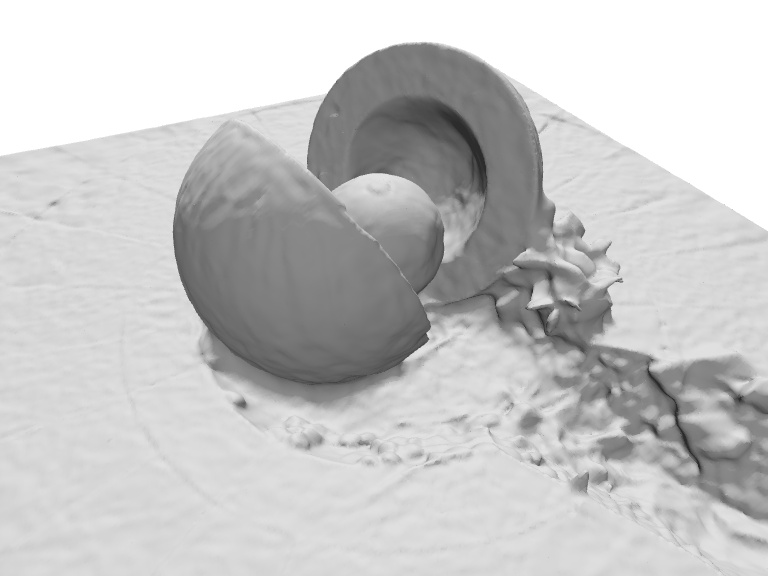}}
  \mpage{0.22}{\includegraphics[width=\linewidth]{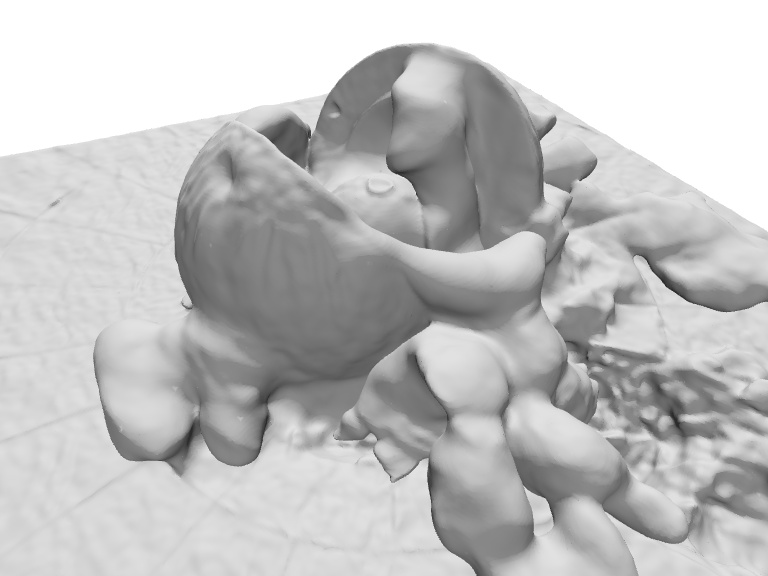}}
  \mpage{0.22}{\includegraphics[width=\linewidth]{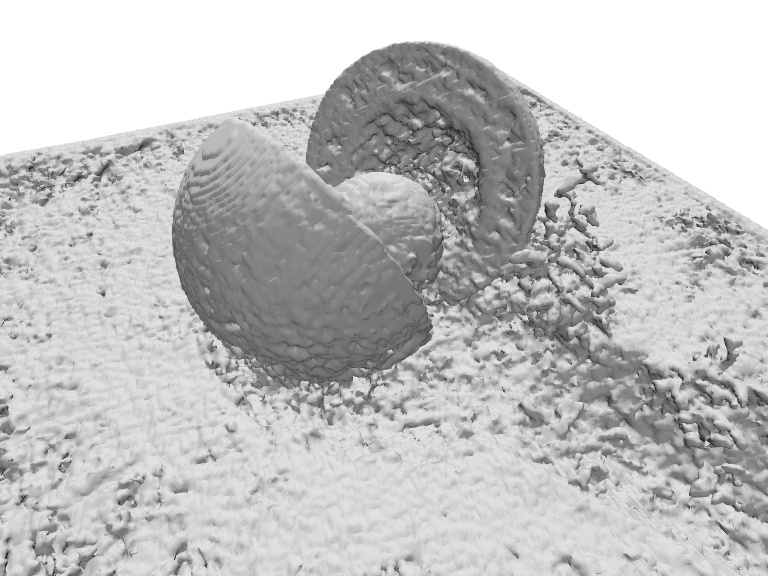}}
  \\
  \mpage{0.05}{\ }
  \mpage{0.215}{Reference Image}
  \mpage{0.215}{Ours}
  \mpage{0.215}{IDR}
  \mpage{0.215}{NeRF}
  \caption{Comparions on surface reconstruction with mask supervision.}
  \label{fig:compare_w_mask}
\end{figure}

\section{Experiments}
\subsection{Experimental settings}\label{sec:experimental_setting}
\heading{Datasets.} To evaluate our approach and baseline methods, we use 15 scenes from the DTU dataset~\cite{DTU2014}, same as those used in IDR~\cite{yariv2020multiview}, with a wide variety of materials, appearance and geometry, including challenging cases for reconstruction algorithms, such as non-Lambertian surfaces and thin structures. Each scene contains 49 or 64 images with the image resolution of $1600\times 1200$. Each scene was tested with and without foreground masks provided by IDR~\cite{yariv2020multiview}.
We further tested on 7 challenging scenes from the low-res set of the BlendedMVS dataset~\cite{yao2020blendedmvs}(CC-4 License). Each scene has $31-143$ images at $768\times 576$ pixels and masks are provided by the BlendedMVS dataset. We further captured two thin objects with 32 input images to test our approach on thin structure reconstruction. 

\heading{Baselines.} (1) The state-of-the-art surface rendering approach --  IDR~\cite{yariv2020multiview}: IDR can reconstruct surface with high quality but requires foreground masks as supervision; Since IDR has demonstrated superior quality compared to another surface rendering based method -- DVR~\cite{niemeyer2020differentiable}, we did not conduct a comparison with DVR.
(2) The state-of-the-art volume rendering approach -- NeRF~\cite{mildenhall2020nerf}: We use a threshold of 25 to extract mesh from the learned density field. We validate this choice in the supplementary material. 
(3) A widely-used classical MVS method -- COLMAP~\cite{schonberger2016pixelwise}: We reconstruct a mesh from the output point cloud of COLMAP with Screened Poisson Surface Reconstruction~\cite{poisson}.
(4) The concurrent work which unifies surface rendering and volume rendering with an occupancy field as scene representation -- UNISURF~\cite{oechsle2021unisurf}.
More details of the baseline methods are included in the supplementary material.

\begin{wraptable}{r}{0.5\linewidth}
    \centering
    \resizebox{\linewidth}{!}{
    \begin{threeparttable}[b]
        \begin{tabular}{c||c|c|c||c|c|c|c}
            \multicolumn{1}{c||}{}&\multicolumn{3}{c||}{\textbf{w/} mask}&\multicolumn{4}{c}{\textbf{w/o} mask} \\
            \hline
            ScanID & IDR & NeRF & Ours & COLMAP & NeRF & UNISURF & Ours \\
            \hline
            scan24 & 1.63 & 1.83 & \bf{0.83} & \bf{0.81} & 1.90 & 1.32 & 1.00 \\
            scan37 & 1.87 & 2.39 & \bf{0.98} & 2.05 & 1.60 & \bf{1.36} & 1.37 \\
            scan40 & 0.63 & 1.79 & \bf{0.56} & \bf{0.73} & 1.85 & 1.72 & 0.93 \\
            scan55 & 0.48 & 0.66 & \bf{0.37} & 1.22 & 0.58 & 0.44 & \bf{0.43} \\
            scan63 & \bf{1.04} & 1.79 & 1.13 & 1.79 & 2.28 & 1.35 & \bf{1.10} \\
            scan65 & 0.79 & 1.44 & \bf{0.59} & 1.58 & 1.27 & 0.79 & \bf{0.65} \\
            scan69 & 0.77 & 1.50 & \bf{0.60} & 1.02 & 1.47 & 0.80 & \bf{0.57} \\
            scan83 & 1.33 & \bf{1.20} & 1.45 & 3.05 & 1.67 & 1.49 & \bf{1.48} \\
            scan97 & 1.16 & 1.96 & \bf{0.95} & 1.40 & 2.05 & 1.37 & \bf{1.09} \\
            scan105 & \bf{0.76} & 1.27 & 0.78 & 2.05 & 1.07 & 0.89 & \bf{0.83} \\
            scan106 & 0.67 & 1.44 & \bf{0.52} & 1.00 & 0.88 & 0.59 & \bf{0.52} \\
            scan110 & \bf{0.90} & 2.61 & 1.43 & 1.32 & 2.53 & 1.47 & \bf{1.20} \\
            scan114 & 0.42 & 1.04 & \bf{0.36} & 0.49 & 1.06 & 0.46 & \bf{0.35} \\
            scan118 & 0.51 & 1.13 & \bf{0.45} & 0.78 & 1.15 & 0.59 & \bf{0.49} \\
            scan122 & 0.53 & 0.99 & \bf{0.45} & 1.17 & 0.96 & 0.62 & \bf{0.54} \\
            \hline
            mean & 0.90 & 1.54 & \bf{0.77} & 1.36 & 1.49 & 1.02 & \bf{0.84} \\

        \end{tabular}
    
    \end{threeparttable}
    }
    \caption{Quantitative evaluation on DTU dataset. COLMAP results are achieved by trim=0.}
    \label{tbl:evaluate_comparision}
    \vspace{-15pt}
\end{wraptable}

\heading{Implementation details.}
We assume the region of interest is inside a unit sphere. We sample 512 rays per batch and train our model for 300k iterations for 14 hours (for the `w/ mask' setting) and 16 hours (for the `w/o mask' setting) on a single NVIDIA RTX2080Ti GPU. For the `w/o mask' setting, we model the background by NeRF++~\cite{zhang2020nerf++}. Our network architecture and initialization scheme are similar to those of IDR~\cite{yariv2020multiview}. More details of the network architecture and training parameters can be found in the supplementary material.

\subsection{Comparisons}
\label{sec:comparision_dtu}
We conducted the comparisons in two settings, with mask supervision (w/ mask) and without mask supervision (w/o mask). We measure the reconstruction quality with the Chamfer distances in the same way as UNISURF~\cite{oechsle2021unisurf} and IDR~\cite{yariv2020multiview} and report the scores in Table~\ref{tbl:evaluate_comparision}. The results show that our approach outperforms the baseline methods on the DTU dataset in both settings -- w/ and w/o mask in terms of the Chamfer distance. Note that the reported scores of IDR in the setting of w/ mask and NeRF and UNISURF in the w/o mask setting are from IDR~\cite{yariv2020multiview} and UNISURF~\cite{oechsle2021unisurf}.

\begin{figure}[!t]
\rotatebox[origin=C]{90}{\parbox{20mm}{\centering \small Scan 69 \\ (DTU)}}
  \mpage{0.22}{\includegraphics[width=\linewidth]{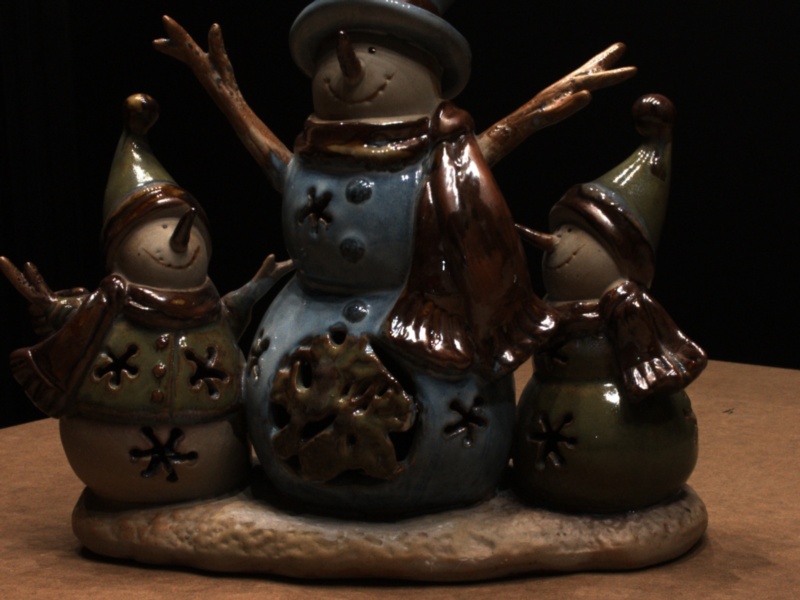}}
  \mpage{0.22}{\includegraphics[width=\linewidth]{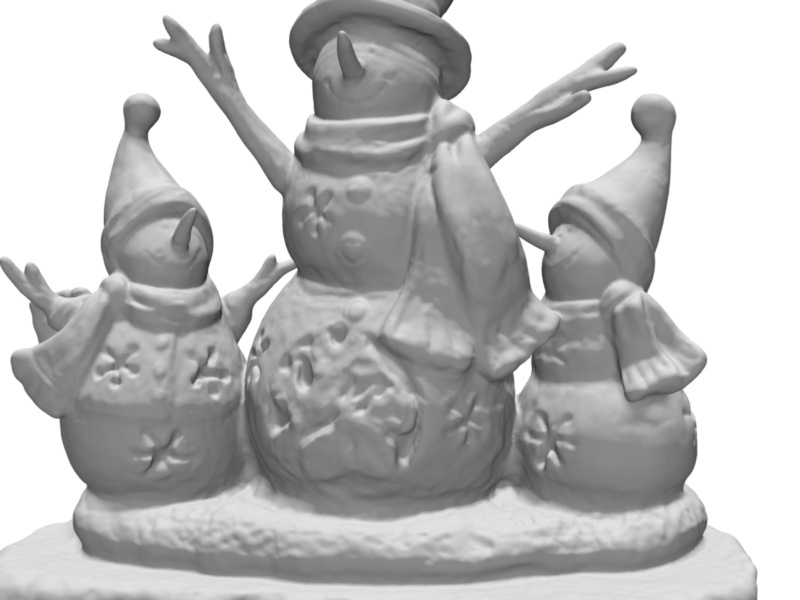}}
  \mpage{0.22}{\includegraphics[width=\linewidth]{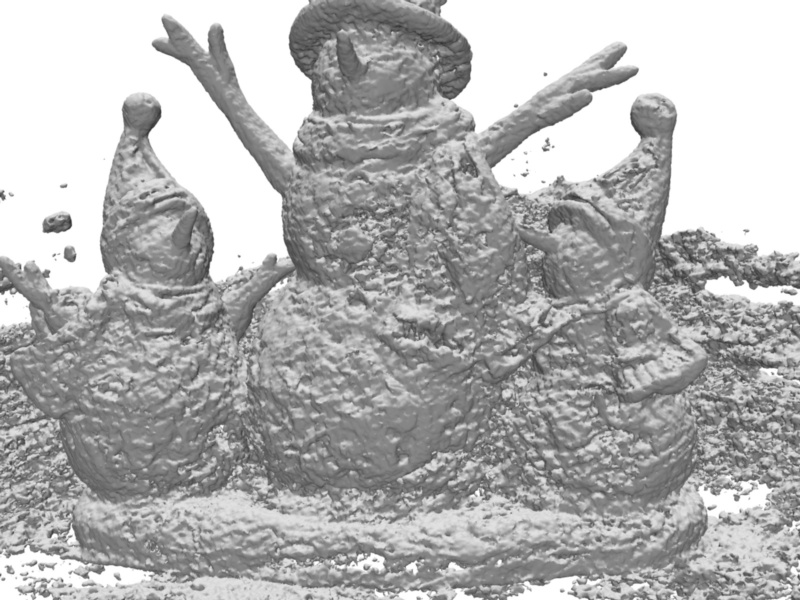}}
  \mpage{0.22}{\includegraphics[width=\linewidth]{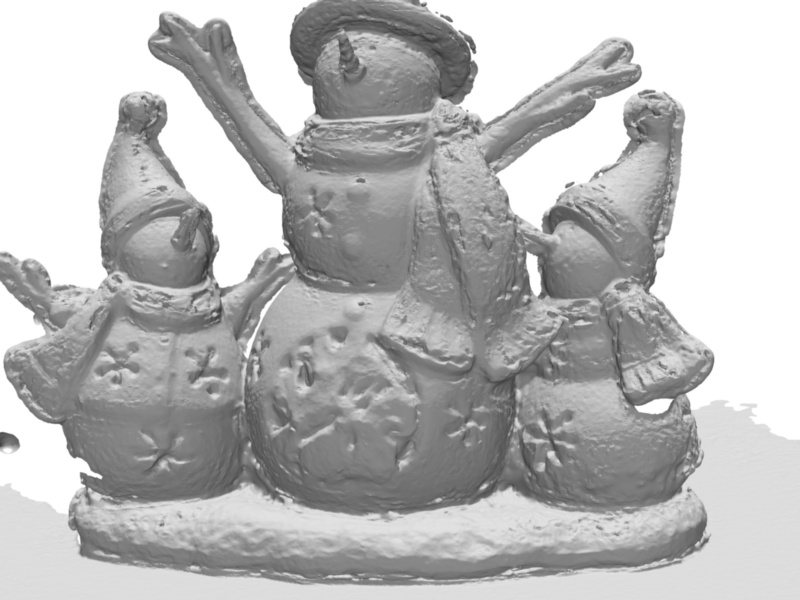}}
 \\
\rotatebox[origin=C]{90}{\parbox{20mm}{\centering \small Clock \\ (BlendedMVS)}} 
  \mpage{0.22}{\includegraphics[width=\linewidth]{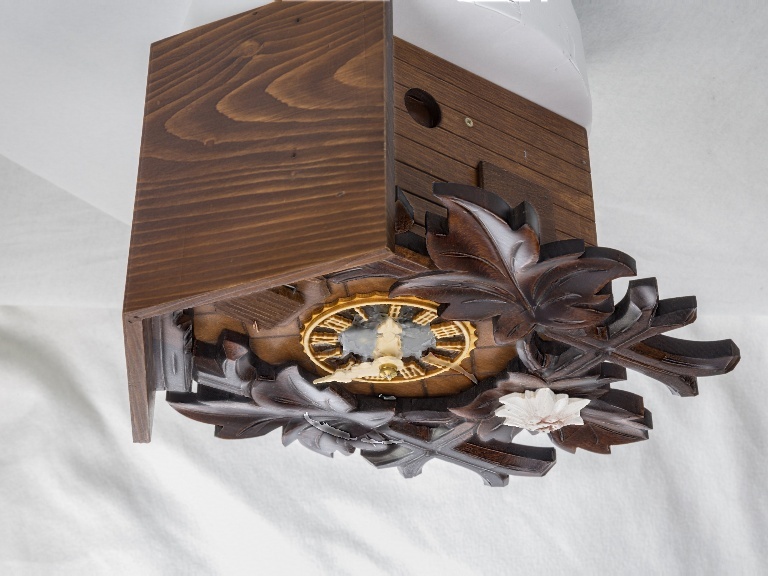}}
  \mpage{0.22}{\includegraphics[width=\linewidth]{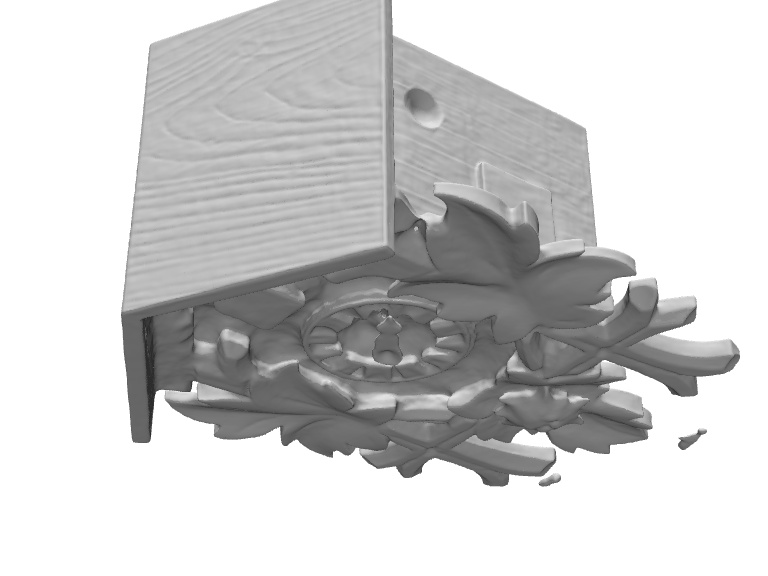}}
  \mpage{0.22}{\includegraphics[width=\linewidth]{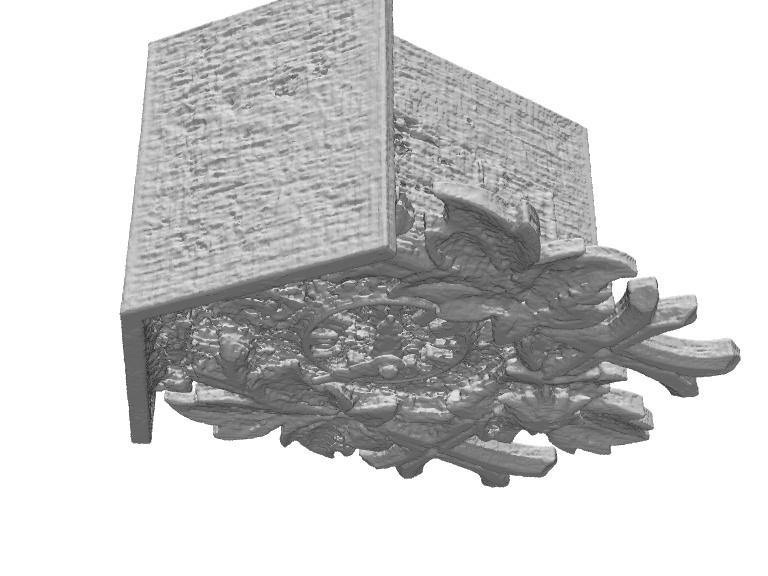}}
  \mpage{0.22}{\includegraphics[width=\linewidth]{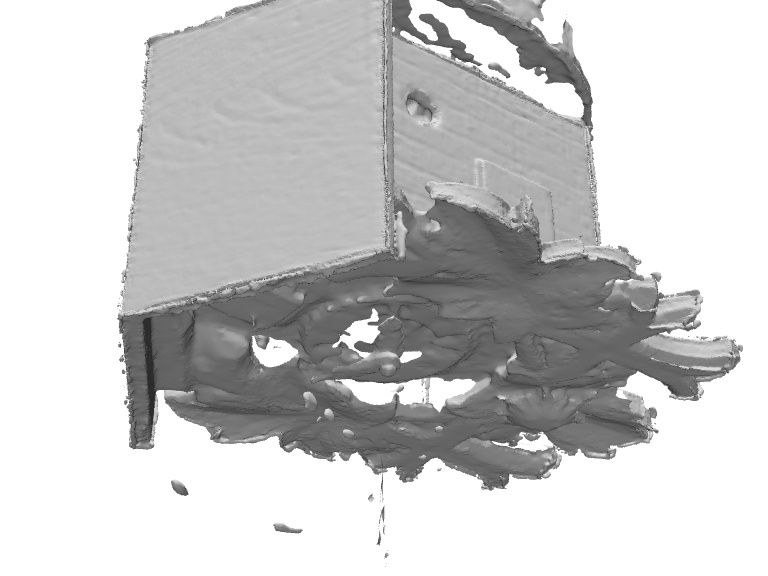}}
  \\
\rotatebox[origin=C]{90}{\parbox{20mm}{\centering \small Sculpture \\ (BlendedMVS)}} 
  \mpage{0.22}{\includegraphics[width=\linewidth]{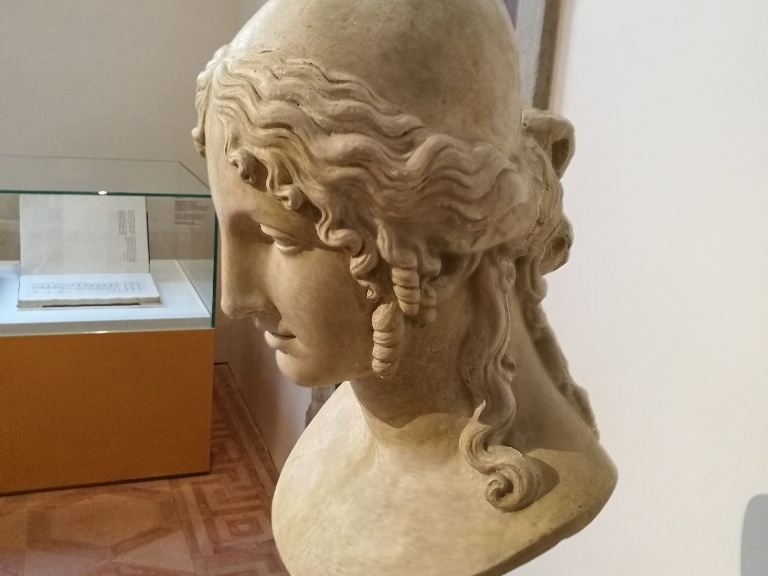}}
  \mpage{0.22}{\includegraphics[width=\linewidth]{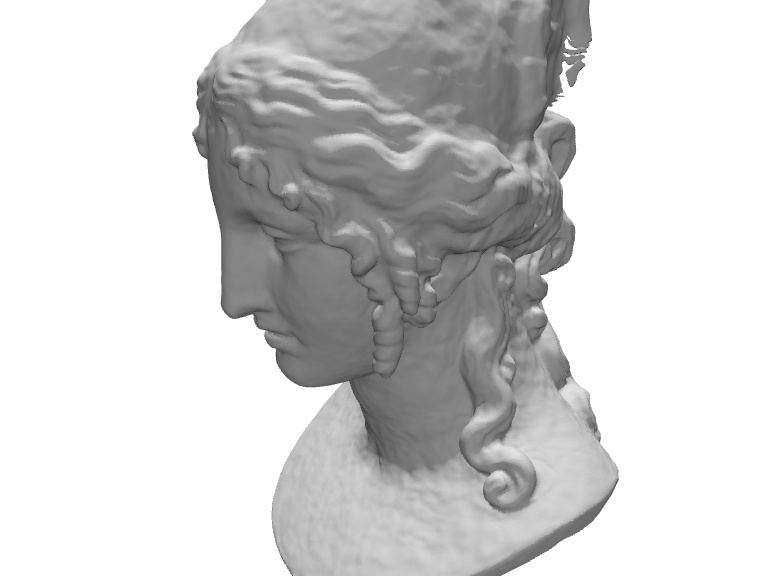}}
  \mpage{0.22}{\includegraphics[width=\linewidth]{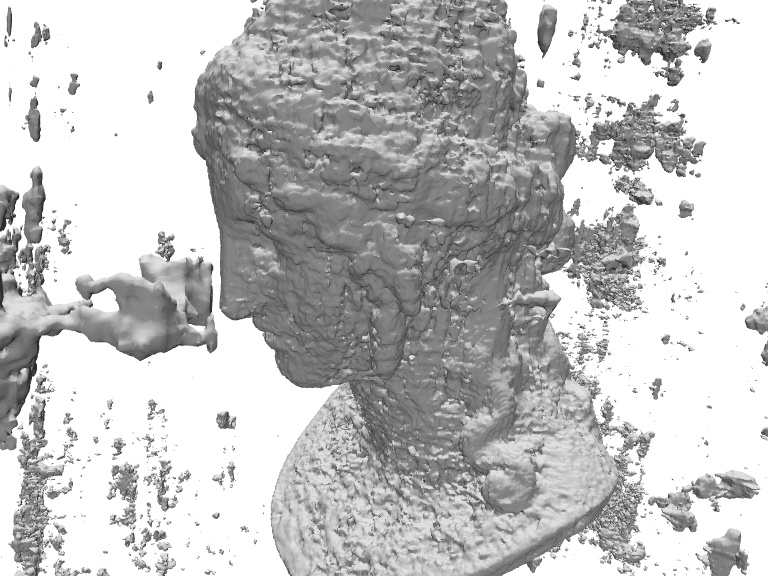}}
  \mpage{0.22}{\includegraphics[width=\linewidth]{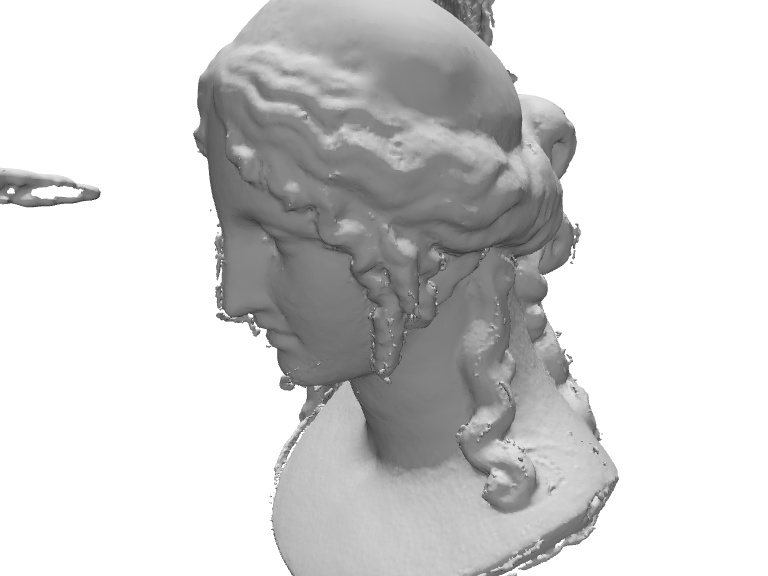}}
  \\
  \rotatebox[origin=C]{90}{\parbox{20mm}{\centering \small Bear \\ (BlendedMVS)}} 
  \mpage{0.22}{\includegraphics[width=\linewidth]{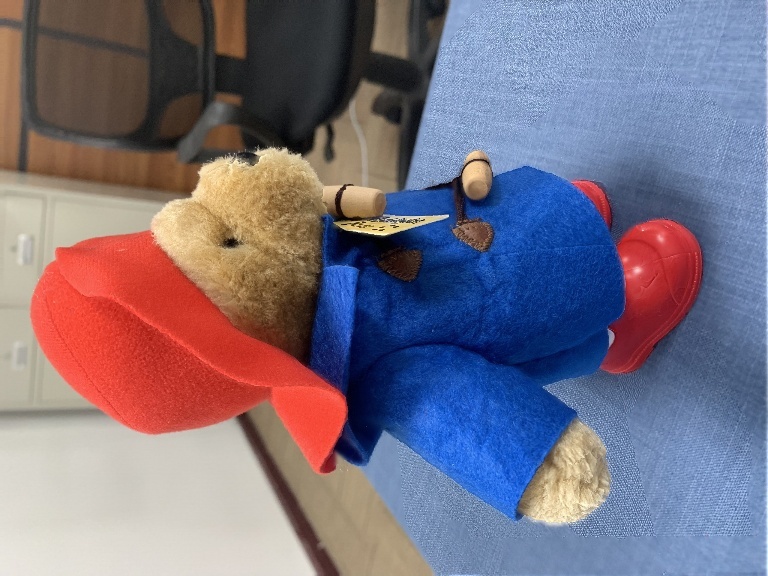}}
  \mpage{0.22}{\includegraphics[width=\linewidth]{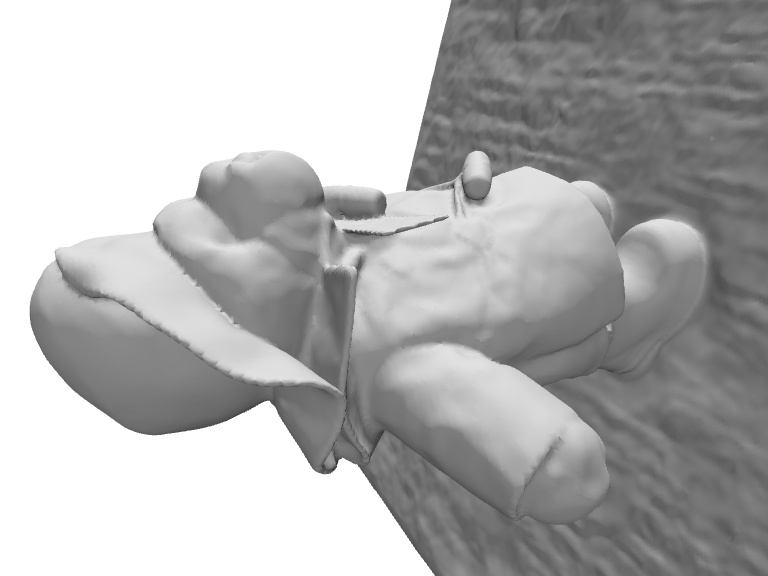}}
  \mpage{0.22}{\includegraphics[width=\linewidth]{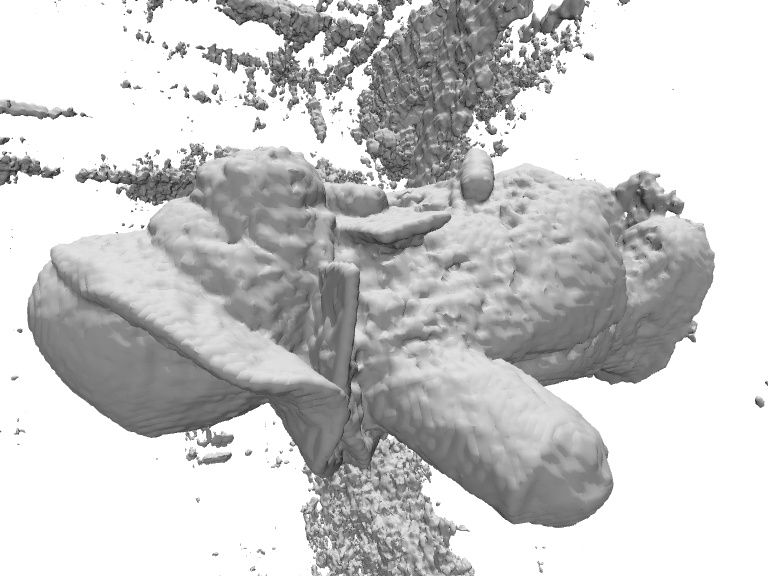}}
  \mpage{0.22}{\includegraphics[width=\linewidth]{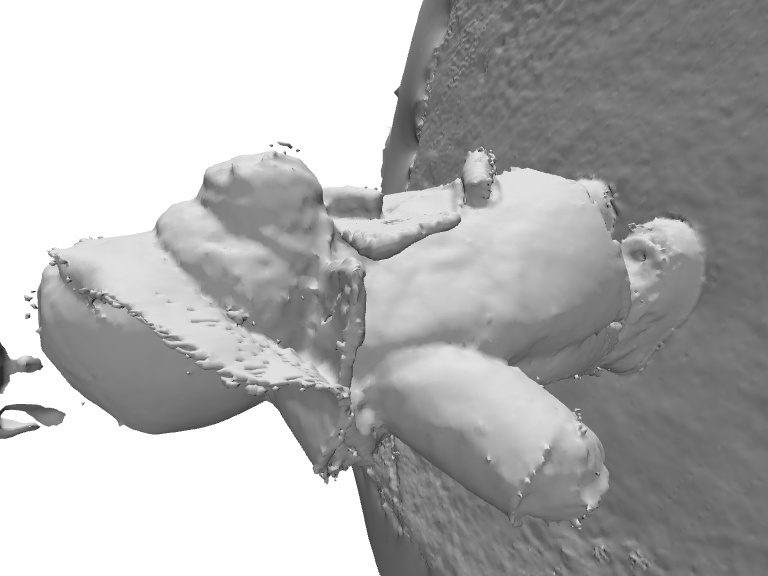}}
  \\
  \mpage{0.05}{\ }
  \mpage{0.215}{Reference Image}
  \mpage{0.215}{Ours}
  \mpage{0.215}{NeRF}
  \mpage{0.215}{COLMAP}
  \caption{Comparions on surface reconstruction without mask supervision.}
  \label{fig:compare_wo_mask}
\end{figure}

We conduct the qualitative comparisons on the DTU dataset and the BlendedMVS dataset in both settings, w/ mask and w/o mask, in \figref{compare_w_mask} and \figref{compare_wo_mask}, respectively.
As shown in \figref{compare_w_mask} for the setting of w/ mask, IDR shows limited performance for reconstructing thin metals parts in Scan 37 (DTU), and fails to handle sudden depth changes in Stone (BlendedMVS) due to the local optimization process in surface rendering. The extracted meshes of NeRF are noisy since the volume density field has not sufficient constraint on its 3D geometry. Regarding the w/o mask setting, we visually compare our method with NeRF and COLMAP in the setting of w/o mask in \figref{compare_wo_mask}, which shows our reconstructed surfaces are with more fidelity than baselines.
We further show a comparison with UNISURF~\cite{oechsle2021unisurf} on two examples in the w/o mask setting. Note that we use the qualitative results of UNISURF reported their paper for comparison. Our method works better for the objects with abrupt depth changes.
More qualitative images are included in the supplementary material.

\begin{figure}[!b]
  \includegraphics[width=\linewidth]{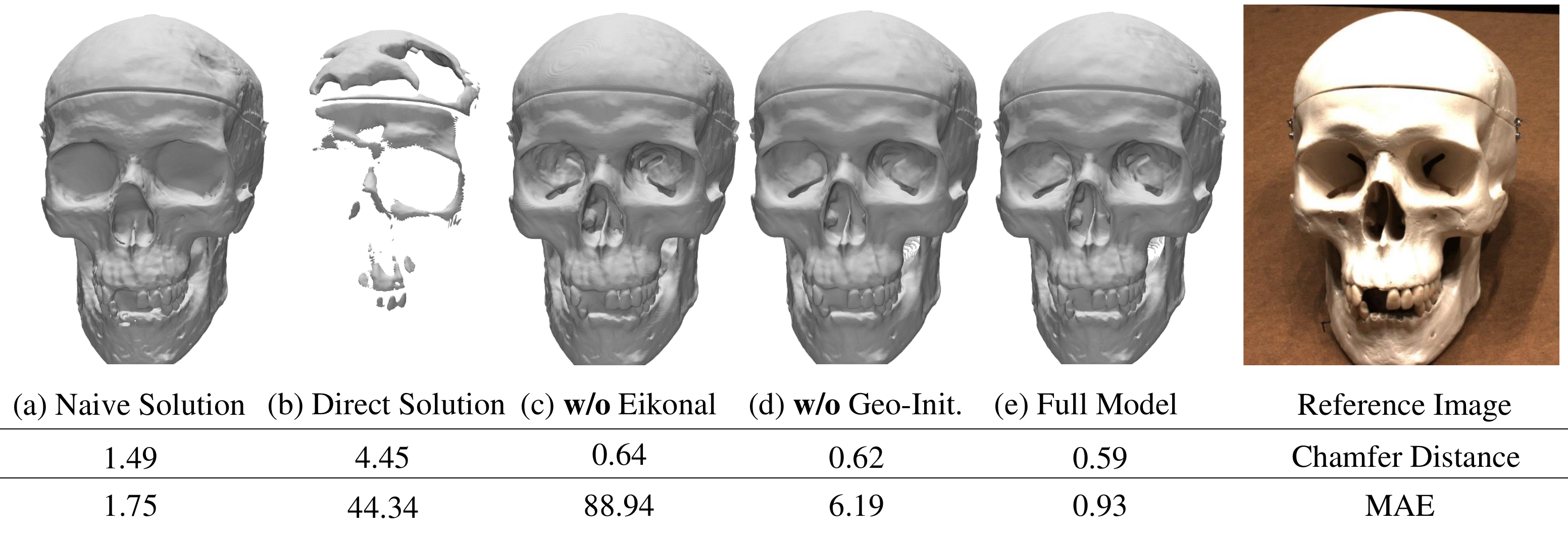}
  \caption{Ablation studies.
  We show the qualitative results and report the quantitative metrics in Chamfer distance and MAE~(mean absolute error) between the ground-truth and predicted SDF values.}
  \label{fig:ablation}
\end{figure}

\subsection{Analysis}
\heading{Ablation study.}
\begin{wrapfigure}{R}{0.5\textwidth}
    \centering
    \includegraphics[width=\linewidth]{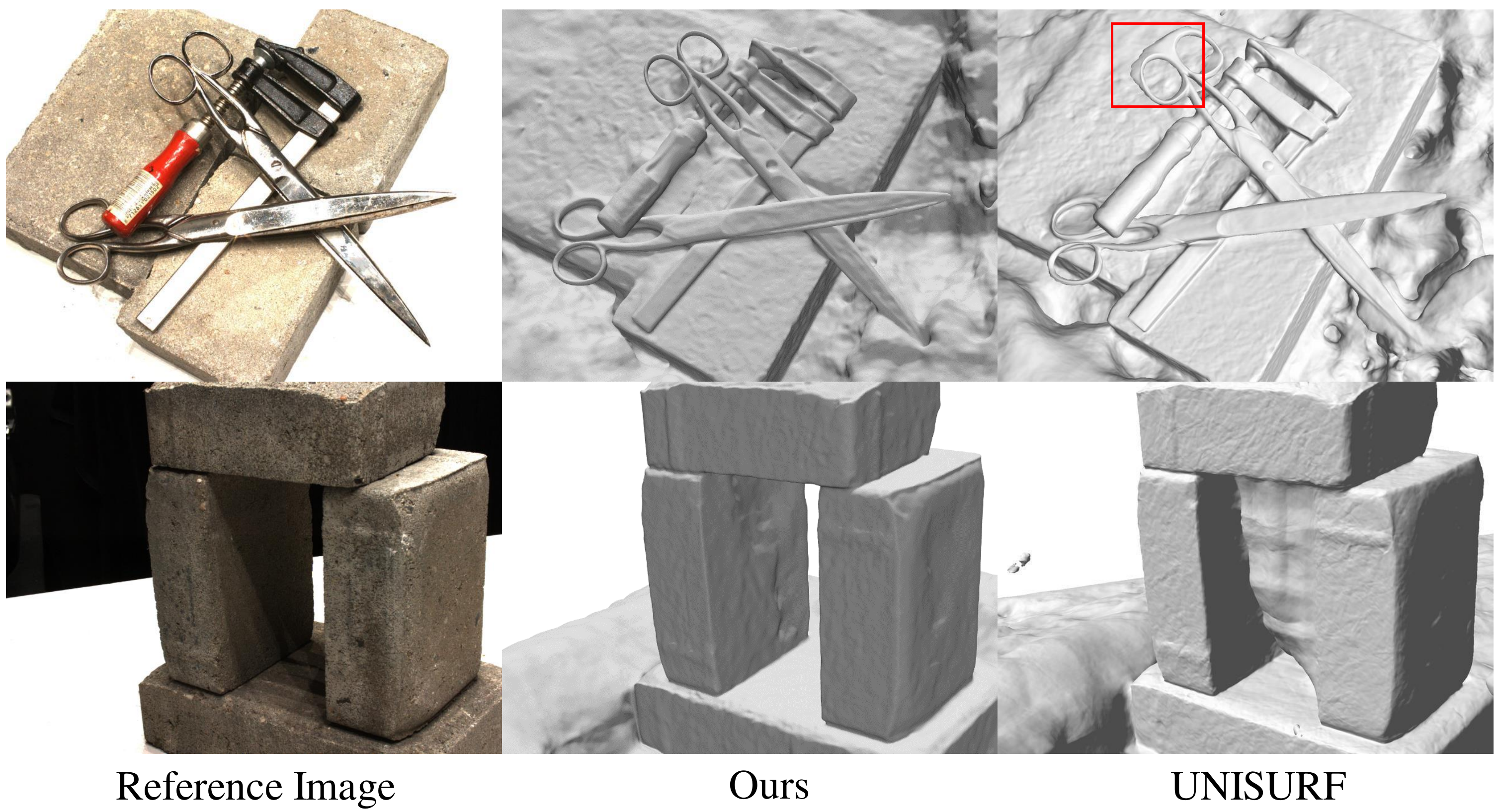}
    \caption{Visual comparisons with UNISURF.}
    \vspace{-10pt}
\end{wrapfigure}

To evaluate the effect of the weight calculation, we test three different kinds of weight constructions described in Sec.~\ref{sec:architecture}: (a) Naive Solution. (b) Straightforward Construction as shown in Eqn.~\ref{eq:trivial1}. (e) Full Model. As shown in \figref{ablation}, the quantitative result of naive solution is worse than our weight choice (e) in terms of the Chamfer distance. This is because it introduces a bias to the surface reconstruction. If direct construction is used, there are severe artifacts.

We also studied the effect of Eikonal regularization~\cite{gropp2020implicit} and geometric initialization~\cite{atzmon2020sal}. Without Eikonal regularization or geometric initialization, the result on Chamfer distance is on par with that of the full model. However, neither of them can correctly output a signed distance function. This is indicated by the MAE(mean absolute error) between the SDF predictions and corresponding ground-truth SDF, as shown in the bottom line of \figref{ablation}. The MAE is computed on uniformly-sampled points in the object's bounding sphere. Qualitative results of SDF predictions are provided in the supplementary material.

\begin{figure}[htb]

  \includegraphics[width=\linewidth]{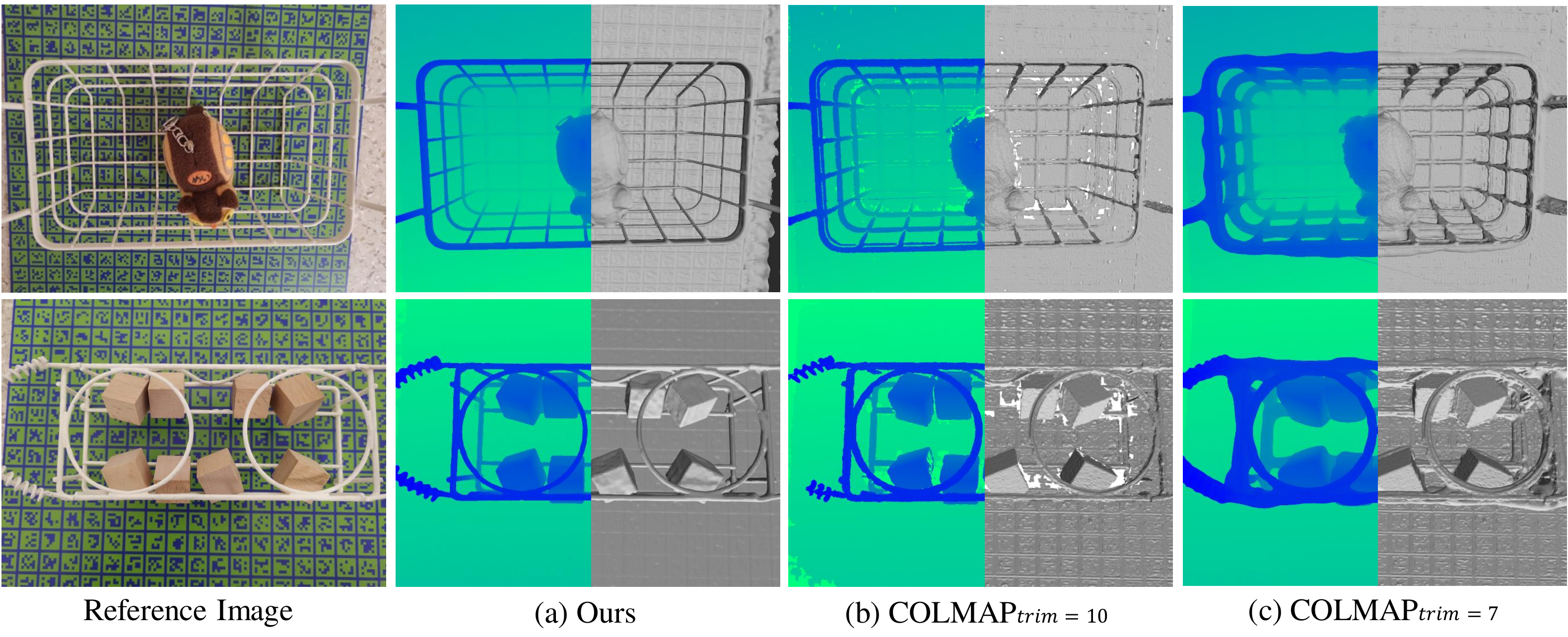}
  \caption{Comparison on scenes with thin structure objects. Left half is the depth map while right half is the reconstructed surface.}
  \label{fig:occ_boundary}
\end{figure}

\heading{Thin structures.}
We additionally show results on two challenging thin objects with 32 input images. The plane with rich texture under the object is used for camera calibration. As shown in Fig.~\ref{fig:occ_boundary}, our method is able to accurately reconstruct these thin structures, especially on the edges with abrupt depth changes. Furthermore, different from the methods \cite{Tabb:cvpr:2013,Liu:sigg:2017,vid2curve,Liu:sigga:2018} which only target at high-quality thin structure reconstruction, our method can handle the scenes which have a mixture of thin structures and general objects. 

\section{Conclusion}
\label{sec:conclusion}
We have proposed {\it NeuS}, a new approach to multiview surface reconstruction that represents 3D surfaces as neural SDF and developed a new volume rendering method for training the implicit SDF representation.
NeuS produces high-quality reconstruction and successfully reconstructs objects with severe occlusions and complex structures. It outperforms the state-of-the-arts both qualitatively and quantitatively.
One limitation of our method is that although our method does not heavily rely on correspondence matching of texture features, the performance would still degrade for textureless objects (we show the failure cases in the supplementary material). Moreover, NeuS has only a single scale parameter $s$ that is used to model the standard deviation of the probability distribution for all the spatial location. Hence, an interesting future research topic is to model the probability with different variances for different spatial locations together with the optimization of scene representation, depending on different local geometric characteristics. Negative societal impact: like many other learning-based works, our method requires a large amount of computational resources for network training, which can be a concern for global climate change.

\section*{Acknowlegements}
We thank Michael Oechsle for providing the results of UNISURF. Christian Theobalt was supported by ERC Consolidator Grant 770784. Lingjie Liu was supported by Lise Meitner Postdoctoral Fellowship. Computational resources are mainly provided by HKU GPU Farm.

\bibliographystyle{plain}
\bibliography{ref}

\setcounter{section}{0}
\renewcommand{\thesection}{\Alph{section}} 
\newpage
\begin{center}
    {\bf\LARGE - Supplementary -}
\end{center}
\vspace{20pt}

\section{Derivation for Computing Opacity $\alpha_i$}

In this section we will derive the formula in Eqn.~13 of the paper for computing the discrete opacity $\alpha_i$. 
Recall that the opaque density function $\rho (t)$ is defined as 
\begin{equation}
\begin{split}
\rho(t) =& \max\left(\frac{-\frac{{\rm d}\Phi_s}{{\rm d} t}(f(\mathbf{p}(t)))}{\Phi_s(f(\mathbf{p}(t)))}, 0\right)\\
=& \max\left(\frac{-(\nabla f(\mathbf{p}(t))\cdot \mathbf{v})\phi_s(f(\mathbf{p}(t)))}{\Phi_s(f(\mathbf{p}(t)))}, 0\right),
\end{split}
\label{eq:rho}
\end{equation}
where $\phi_s(x)$ and $\Phi_s(x)$ are the probability density function (PDF) and cumulative distribution function (CDF) of logistic distribution, respectively.
First consider the case where the sample point interval $[t_i, t_{i+1}]$ lies in a range $[t_\ell, t_r]$ over which the camera ray is entering the surface from outside to inside, i.e. the signed distance function is decreasing on the camera ray $\mathbf{p}(t)$ over $[t_\ell, t_r]$. Then it is easy to see that $-(\nabla f(\mathbf{p}(t))\cdot \mathbf{v}) > 0 $ in $[t_i, t_{i+1}]$. It follows from Eqn.~12 of the paper that,
\begin{equation}
\begin{split}
\alpha_i =& 1 - \exp\left(-\int_{t_i}^{t_{i+1}}\rho(t){\rm d}t\right)\\
=& 1 - \exp\left(-\int_{t_i}^{t_{i+1}}\frac{-(\nabla f(\mathbf{p}(t))\cdot \mathbf{v})\phi_s(f(\mathbf{p}(t)))}{\Phi_s(f(\mathbf{p}(t)))}{\rm d}t\right).
\end{split}
\label{eq:alpha}
\end{equation}
Note that the integral term is computed by
\begin{equation}
\int \frac{-(\nabla f(\mathbf{p}(t))\cdot \mathbf{v})\phi_s(f(\mathbf{p}(t)))}{\Phi_s(f(\mathbf{p}(t)))}{\rm d}t = -\ln(\Phi_s(f(\mathbf{p}(t)))) + C,
\end{equation}
where $C$ is a constant. Thus the discrete opacity can be computed by
\begin{equation}
\begin{split}
\alpha_i=& 1 - \exp\left[-\left(-\ln(\Phi_s(f(\mathbf{p}(t_{i+1})))) + \ln(\Phi_s(f(\mathbf{p}(t_{i}))))\right)\right]\\
=& 1 - \frac{\Phi_s(f(\mathbf{p}(t_{i+1})))}{\Phi_s(f(\mathbf{p}(t_{i})))}\\
=& \frac{\Phi_s(f(\mathbf{p}(t_{i}))) - \Phi_s(f(\mathbf{p}(t_{i+1})))}{\Phi_s(f(\mathbf{p}(t_i)))}.
\end{split}
\end{equation}

Next consider the case where $[t_i, t_{i+1}]$ lies in a range $[t_\ell, t_r]$ over which the camera ray is exiting the surface, i.e. the signed distance function is increasing on $\mathbf{p}(t)$ over $[t_\ell, t_r]$. Then we have $-(\nabla f(\mathbf{p}(t))\cdot \mathbf{v}) < 0$ in $[t_i, t_{i+1}]$. Then, according to Eqn.~\ref{eq:rho}, we have $\rho(t) = 0$. Therefore, by Eqn.~12 of the paper, we have 
\[
\alpha_i = 1 - \exp\left(-\int_{t_i}^{t_{i+1}}\rho(t){\rm d}t\right) = 1 - \exp\left(-\int_{t_i}^{t_{i+1}} 0 {\rm d}t\right) = 0.
\]
Hence, the alpha value $\alpha_i$ in this case is given by
\begin{equation}
\begin{split}
\alpha_i = \max \left(\frac{\Phi_s(f(\mathbf{p}(t_{i}))) - \Phi_s(f(\mathbf{p}(t_{i+1})))}{\Phi_s(f(\mathbf{p}(t_i)))}, 0\right).
\end{split}
\end{equation}
This completes the derivation of Eqn.~13 of the paper.

\section{First-order Bias Analysis}
\label{sec:first_order}
\subsection{Proof of Unbiased Property of Our Solution}
\label{sec:proof}
{\sc Proof of Theorem 1}:
Suppose that the ray is going from outside to inside of the surface. Hence, we have $-(\nabla f(\mathbf{p}(t))\cdot \mathbf{v}) > 0$, because by convention the signed distance function $f(\mathbf{x})$ is positive outside and negative inside of the surface. 

Recall that our S-density field $\phi_s(f(\mathbf{x}))$ is defined using the logistic density function $\phi_s(x) = se^{-sx}/(1 + e^{-sx})^2$, which is the derivative of the Sigmoid function $\Phi_s(x) = (1 + e^{-sx})^{-1}$, i.e. $\phi_s(x) = \Phi_s'(x)$. 

According to Eqn.~5 of the paper, the weight function $w(t)$ is given by
\[
    w(t) = T(t)\rho(t),
\]
where  
\[
\rho (t) = \max\left(\frac{-(\nabla f(\mathbf{p}(t))\cdot \mathbf{v})\phi_s(f(\mathbf{p}(t)))}{ \Phi_s(f(\mathbf{p}(t)))}, 0\right).
\]

By assumption, $-(\nabla f(\mathbf{p}(t))\cdot \mathbf{v}) > 0$ for $t \in \left[t_l, t_r\right]$. Since $\phi_s$ is a probability density function, we have $\phi_s(f(\mathbf{p}(t))) > 0$. Clearly, $\Phi_s(f(\mathbf{p}(t))) > 0$. It follows that 
\[
\rho(t) = \frac{-(\nabla f(\mathbf{p}(t))\cdot \mathbf{v})\phi_s(f(\mathbf{p}(t)))}{\Phi_s(f(\mathbf{p}(t)))},
\]
which is positive. Hence,
\begin{equation}
\begin{split}    
    w(t) =& T(t)\rho(t)\\
    =&\exp\left(-\int_{0}^{t} \rho(t'){\rm d}t'\right)\rho(t)\\
    =&\exp\left(-\int_{0}^{t_l} \rho(t'){\rm d}t'\right)\exp\left(-\int_{t_l}^{t} \rho(t'){\rm d}t'\right)\rho(t)\\
    =&T(t_l)\exp\left(-\int_{t_l}^{t} \rho(t'){\rm d}t'\right)\rho(t) \\
    =&T(t_l)\exp\left[-(-\ln(\Phi_s(f(\mathbf{p}(t)))) + \ln(\Phi_s(f(\mathbf{p}(t_{l})))))\right]\rho(t)\\
    =&T(t_l)\frac{\Phi_s(f(\mathbf{p}(t)))}{\Phi_s(f(\mathbf{p}(t_l)))}\frac{-(\nabla f(\mathbf{p}(t))\cdot \mathbf{v})\phi_s(f(\mathbf{p}(t)))}{\Phi_s(f(\mathbf{p}(t)))}\\
    =&\frac{-(\nabla f(\mathbf{p}(t))\cdot \mathbf{v})T(t_l)}{\Phi_s(f(\mathbf{p}(t_l)))}\phi_s(f(\mathbf{p}(t))).
\end{split}
\end{equation}

As a first-order approximation of signed distance function $f$, suppose that locally the surface is tangentially approximated by a sufficiently small planar patch with its outward unit normal vector denoted as $\mathbf{n}$. Because $f(\mathbf{x})$ is a signed distance function, locally it has a unit gradient vector $\nabla f =  \mathbf{n}$. Then we have 
\begin{equation}
\begin{split}
w(t)=&\frac{-(\nabla f(\mathbf{p}(t))\cdot \mathbf{v})T(t_l)}{\Phi_s(f(\mathbf{p}(t_l)))}\phi_s(f(\mathbf{p}(t)))\\
    =&\frac{|\cos (\theta)| T(t_l)}{\Phi_s(f(\mathbf{p}(t_l)))}\phi_s(f(\mathbf{p}(t))),
\end{split}
\end{equation}
where $\theta$ is the angle between the view direction $\mathbf{v}$ and the unit normal vector $\mathbf{n}$, that is, $\cos (\theta) = \mathbf{v}\cdot \mathbf{n}$. Here $|\cos (\theta)| T(t_l)\cdot \Phi_s(f(\mathbf{p}(t_l)))^{-1}$ can be regarded as a constant. Hence, $w(t)$ attains a local maximum when $f(\mathbf{p}(t)) = 0$ because $\phi_s(x)$ is a unimodal density function attaining the maximal value at $x=0$. 

We remark that in this proof we do not make any assumption on the existence of surfaces between the camera and the sample point $\mathbf{p}(t_l)$. Therefore the conclusion holds true for the case of multiple surface intersections on the camera ray. This completes the proof. 
$\Box$

\subsection{Bias in Naive Solution}
In this section we show that the weight function derived in naive solution is biased. According to Eqn.~3 of the paper, $w(t) = T(t)\sigma(t)$, with the opacity $\sigma(t) = \phi_s(f(\mathbf{p}(t)))$.
Then we have
\begin{equation}
\begin{split}
\frac{{\rm d} w}{{\rm d}t} =& \frac{{\rm d} (T(t)\sigma(t))}{{\rm d}t}\\
=& \frac{{\rm d} T(t)}{{\rm d}t} \sigma(t) + T(t)\frac{{\rm d} \sigma(t)}{{\rm d}t}\\
=& \left[\exp\left(-\int_{0}^{t}\sigma(t){\rm d}t\right)(-\sigma(t))\right]\sigma(t) + T(t)\frac{{\rm d} \sigma(t)}{{\rm d}t}\\
=& T(t)(-\sigma(t))\sigma(t) + T(t)\frac{{\rm d} \sigma(t)}{{\rm d}t}\\
=& T(t)\left(\frac{{\rm d} \sigma(t)}{{\rm d}t} - \sigma(t)^2\right).
\end{split}
\label{eq:dwdt}
\end{equation}

Now we perform the same first-order approximation of signed distance function $f$ near the surface intersection as in \secref{proof}. In this condition, the above equation can be rewritten as
\begin{equation}
\begin{split}
\frac{{\rm d} w}{{\rm d} t} =& T(t)\left((\nabla f(\mathbf{p}(t)) \cdot \mathbf{v}) \phi_s'(f(\mathbf{p}(t))) - \phi_s(f(\mathbf{p}(t)))^2\right) \\
=& T(t)\left(\cos(\theta) \phi_s'(f(\mathbf{p}(t))) - \phi_s(f(\mathbf{p}(t)))^2\right).
\end{split}
\end{equation}

Here $\cos(\theta)$ can be regarded as a constant. Now suppose $\mathbf{p}(t^*)$ is a point on the surface $\mathbb{S}$, that is, $f(\mathbf{p}(t^*))=0$. Next we will examine the value of $\frac{{\rm d} w}{{\rm d}t} (t)$ at $t=t^*$. First, clearly, $T(t^*) >0$ and $\phi_s(f(\mathbf{p}(t^*)))^2 >0$. Then, since $\phi_s'(0) = 0$, we have
\[
\frac{{\rm d} w}{{\rm d}t}(t^*) = T(t^*)(\cos(\theta) \phi_s'(0) - \sigma(t^*)^2) = - T(t^*)\phi_s(0)^2 < 0.
\]
Hence $w(t)$ in naive solution does not attain a local maximum at $t=t^*$, which corresponds to a point on the surface $\mathbb{S}$. This completes the proof. $\Box$

\section{Second-order Bias Analysis}

In this section we briefly introduce our local analysis in the interval $[t_l, t_r]$ near the surface intersection, in second-order approximation. In this condition, we follow the similar assumption as \secref{first_order} that the signed distance function $f(\mathbf{p}(t))$ monotonically decreases along the ray in the interval $[t_l, t_r]$.

According to Eqn. \ref{eq:dwdt}, the derivative of $w(t)$ is given by:
\[
\frac{{\rm d} w}{{\rm d}t} = T(t)\left(\frac{{\rm d} \sigma(t)}{{\rm d}t} - \sigma(t)^2\right).
\]

\begin{figure}[hbt]
  \includegraphics[width=\linewidth]{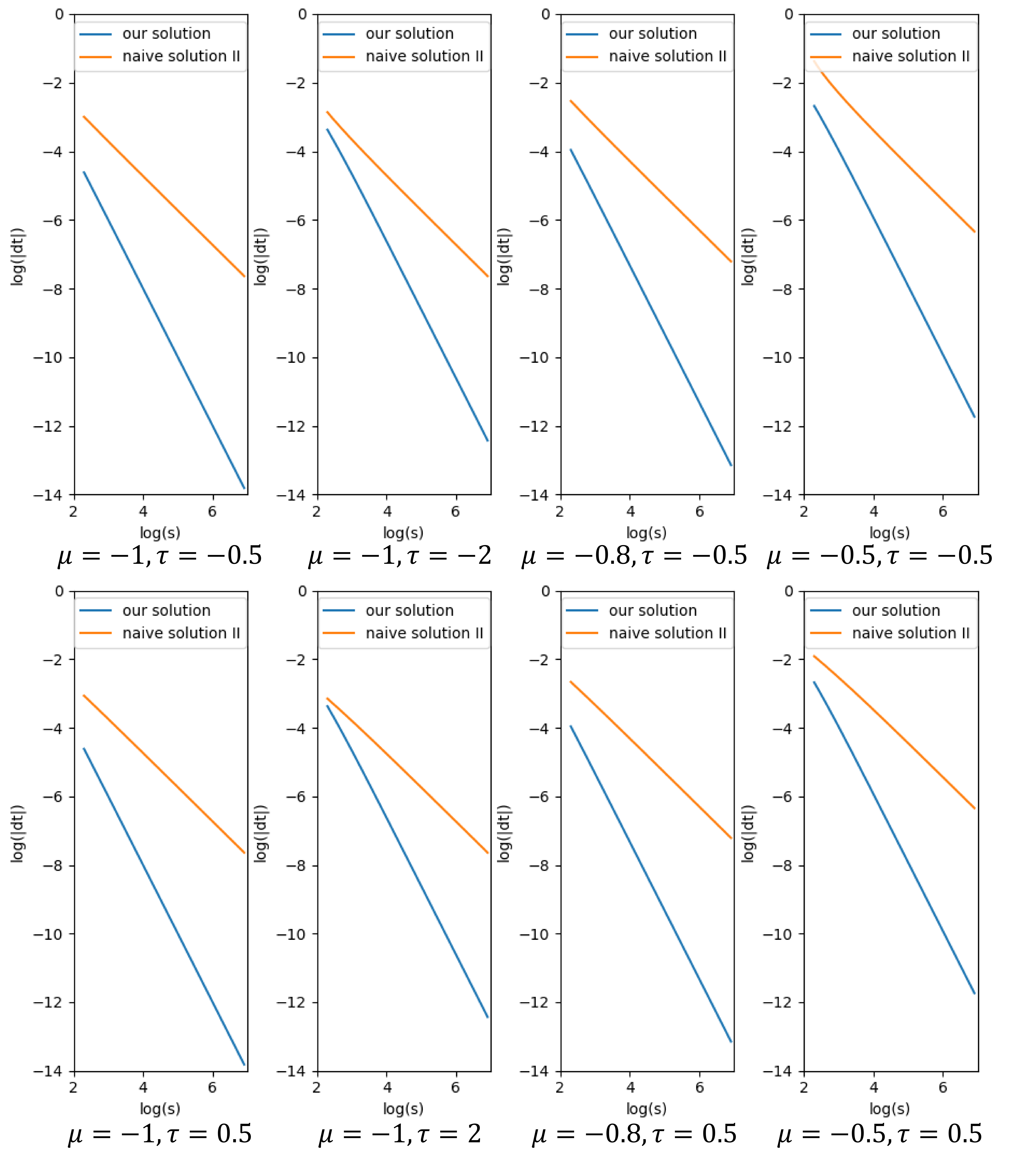}\\
  \caption{The curve of $\Delta_t$ versus $s$, given fixed $\mu, \tau$. Note that the axes are illustrated in $\ln(|\Delta_t|)$ and $\ln(s)$.}
  \label{fig:error}
\end{figure}

Clearly, we have $T(t) > 0$. Hence, when $w(t)$ attains local maximum at $\bar t$, there is $\left(\frac{{\rm d} \sigma(\bar t)}{{\rm d} t} - \sigma(\bar t)^2\right) = 0$.

{\bf The case of our solution}. In our solution, the volume density is given by $\sigma(t) = \rho(t)$ following Eqn. \ref{eq:rho}. After organizing, we have
\[
\frac{{\rm d}^2f}{{\rm d} t}(\mathbf{p}(\bar t))\cdot \phi_s(f(\mathbf{p}(\bar t))) +\left(\frac{{\rm d}{f}}{{\rm d}{t}}(\mathbf{p}(\bar t))\right)^2\phi_s^{'}(f(\mathbf{p}(\bar t))) = 0.
\]
Here we perform a local analysis at $\bar t$ near the surface intersection $t^*$, where $f(\mathbf{p}(t^*))=0$, $\bar t = t^* + \Delta_t$. And we let $\frac{{\rm d} f}{\rm dt}(\mathbf{p}(t^*))=\mu$, and $\frac{{\rm d}^2f}{{\rm d}t^2}(\mathbf{p}(t^*)) = \tau$. As a second-order analysis, we assume that in this local interval $t \in [t_l, t_r]$, $\frac{{\rm d}^2f}{{\rm d}t^2}(\mathbf{p}(t))$ is fixed. After substitution and organization, the induced equation for local maximum point $\bar t$ is
\begin{equation}
\tau \cdot \left(1 + e^{-s(\mu\Delta_t + \frac{1}{2}\tau\Delta_t^2)}\right) = (\mu + \tau\Delta_t)^2\cdot\left(s\left(1 - e^{-s(\mu\Delta_t + \frac{1}{2}\tau\Delta_t^2)}\right)\right),
\label{eq:ours_local}
\end{equation}
which we will analyze later.

{\bf The case of the naive solution}. Here we conduct a similar local analysis as in case of our solution. Regarding naive solution, when $w(t)$ attains local maximum at $\bar t$, there is:
\begin{equation}
    (\mu + \tau\Delta_t) \cdot\left(-\left(1 - e^{-2s(\mu\Delta_t + \frac{1}{2}\tau\Delta_t^2)}\right)\right)=e^{-s(\mu\Delta_t + \frac{1}{2}\tau\Delta_t^2)}.
    \label{eq:naive_local}
\end{equation}

{\bf Comparison.}
Based on Eqn. \ref{eq:ours_local} and Eqn. \ref{eq:naive_local}, we can numerically solve the equations on $\Delta_t$ for any given values of $\mu, \tau,$ and $s$. Below we plot the curves of $\Delta_t$ versus increasing $s$ for different (fixed) values of $\mu, \tau$ in Fig. \ref{fig:error}.

As shown in Fig. \ref{fig:error}, the error of local maximum position $\Delta_t = O(s^{-2})$ for our solution and the error $\Delta_t = O(s^{-1})$ for the naive solution. That is to say, our error converges to zero faster than the error of the naive solution does as the standard deviation $1/s$ of the $S$-density approaches to $0$, which is quadratic convergence versus linear convergence.

\section{Additional Experimental Details}
\subsection{Additional Implemenation Details}
{\bf Network architecture}. We use a similar network architecture as IDR~\cite{yariv2020multiview}, which consists of two MLPs to encode SDF and color respectively. The signed distance function $f$ is modeled by an MLP that consists of 8 hidden layers with hidden size of 256. We replace original ReLU with Softplus with $\beta=100$ as activation functions for all hidden layers. A skip connection~\cite{park2019deepsdf} is used to connect the input with the output of the fourth layer. The function $c$ for color prediction is modeled by a MLP with 4 hidden layers with size of 256, which takes not only the spatial location $\mathbf{p}$ as inputs but also the view direction $\mathbf{v}$, the normal vector of SDF $\mathbf{n}=\nabla{f}(\mathbf{p})$, and a 256-dimensional feature vector from the SDF MLP. Positional encoding is applied to spatial location $\mathbf{p}$ with 6 frequencies and to view direction $\mathbf{v}$ with 4 frequencies. Same as IDR, we use weight normalization~\cite{salimans2016weight} to stabilize the training process.

{\bf Training details}. We train our neural networks using the ADAM optimizer~\cite{kingma2014adam}. The learning rate is first linearly warmed up from 0 to $5\times 10^{-4}$ in the first 5k iterations, and then controlled by the cosine decay schedule to the minimum learning rate of $2.5\times 10^{-5}$. We train each model for 300k iterations for 14 hours (for the `w/ mask' setting) and 16 hours (for the `w/o mask' setting) in total on a single Nvidia 2080Ti GPU.

\begin{figure}[htb]
  \includegraphics[width=\linewidth]{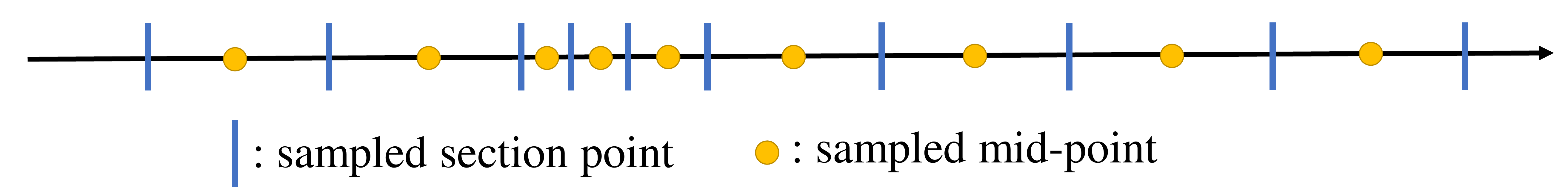}\\
  \caption{The section points and mid-points defined on a ray.}
  \label{fig:method_sample}
\end{figure}

{\bf Alpha and color computation}. In the implementation, we actually have two types of sampling points - the sampled section points $\mathbf{q}_i = \mathbf{o} + t_i\mathbf{v}$ and the sampled mid-points $\mathbf{p}_i = \mathbf{o} + \frac{t_i + t_{i+1}}{2}\mathbf{v}$, with section length $\delta_i=t_{i + 1} - t_{i}$, as illustrated in \figref{method_sample}. 
To compute the alpha value $\alpha_i$, we use the section points, which is $\max(\frac{\Phi_s(f(\mathbf{q}_{i})) - \Phi_s(f(\mathbf{q}_{i+1}))}{\Phi_s(f(\mathbf{q}_{i}))}, 0)$.
To compute the color $c_i$, we use the color of the mid-point $\mathbf{p}_i$.

{\bf Hierarchical sampling}. Specifically, we first uniformly sample 64 points along the ray, then we iteratively conduct importance sampling for $k=4$ times. The coarse probability estimation in the i-th iteration is computed by a fixed $s$ value, which is set as $32\times 2^{i}$. In each iteration, we additionally sample 16 points. Therefore, the total number of sampled points for NeuS is 128. For the `w/o mask' setting, we sample extra 32 points outside the sphere. The outside scene is represented using NeRF++~\cite{zhang2020nerf++}.

\begin{table}[!b]
    \centering
    \begin{tabular}{c|c|c|c|c|c}
         Scan ID & Threshold 0 & Threshold 25 & Threshold 50 & Threshold 100 & Threshold 500 \\
         \hline
         Scan 40 & 2.36 & {\bf 1.79} & 1.86 & 2.07 & 4.26 \\
         Scan 83 & 1.65 & {\bf 1.20} & 1.37 & 2.24 & 29.10 \\
         Scan 114 & 1.62 & {\bf 1.04} & 1.10 & 1.43 & 8.66
    \end{tabular}
    \caption{The Chamfer distances between the ground-truth and the level-set surfaces extracted from the NeRF results using different threshold values on three scenes from the DTU dataset.}
    \label{tbl:nerf_eval}
\end{table}

\subsection{Baselines}
{\bf IDR\cite{yariv2020multiview}}. To implement IDR, we use their officially released codes\footnote{https://github.com/lioryariv/idr} and pretrained models on the DTU dataset. 

\begin{figure}[!t]
  \includegraphics[width=\linewidth]{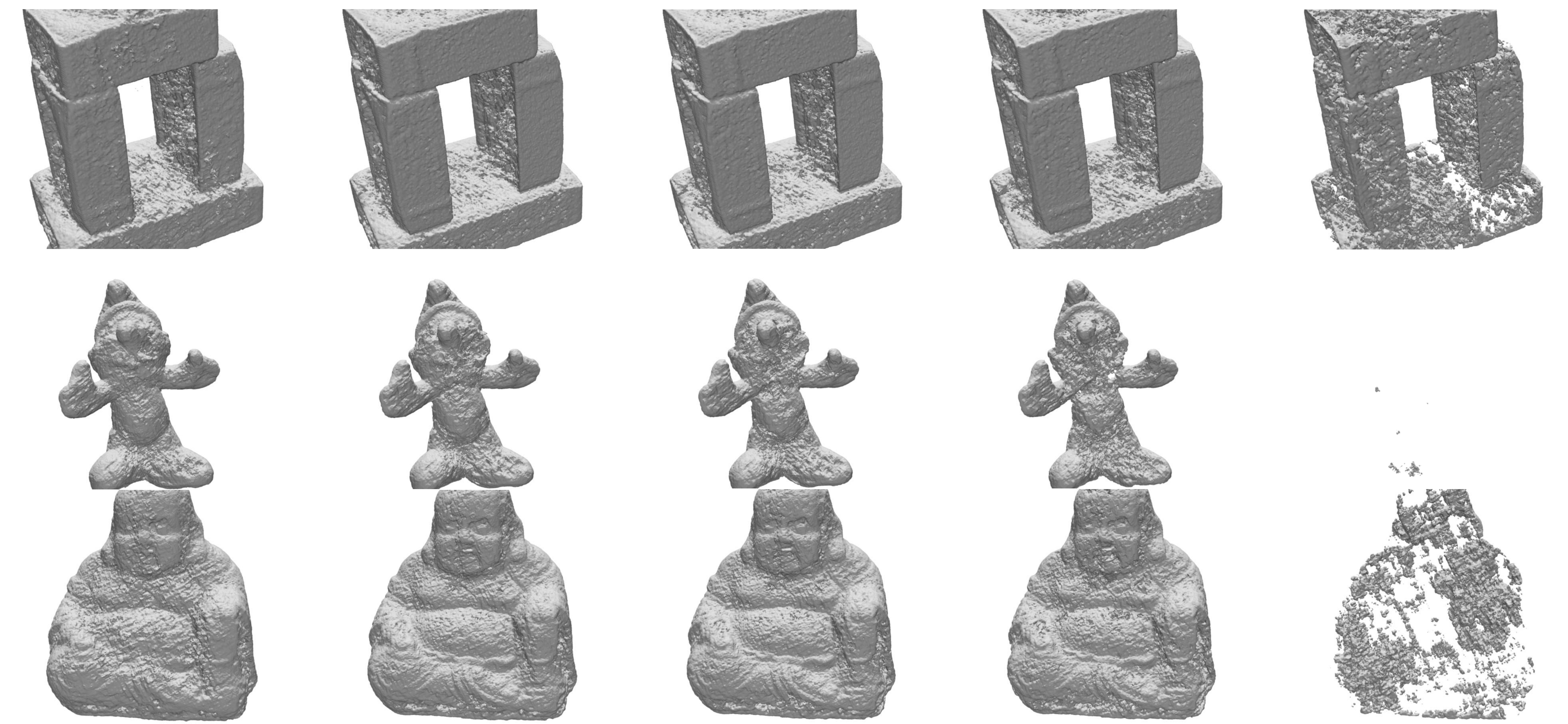}\\
  \mpage{0.1875}{Threshold 0}
  \mpage{0.1875}{Threshold 25}
  \mpage{0.1875}{Threshold 50}
  \mpage{0.1875}{Threshold 100}
  \mpage{0.1875}{Threshold 500}
  \caption{The visualization of the level-set surfaces extracted from the NeRF results using different threshold values.}
  \label{fig:nerf}
\end{figure}

{\bf NeRF\cite{mildenhall2020nerf}}. To implement NeRF, we use the code from nerf-pytorch\footnote{https://github.com/yenchenlin/nerf-pytorch}.
To extract surfaces from NeRF, we use the density level-set of 25, which is validated by experiments to be the best level-set with smallest reconstruction errors, as shown in \tblref{nerf_eval} and \figref{nerf}.

{\bf COLMAP\cite{schonberger2016pixelwise}}. We use the officially provided CLI(command line interface) version of COLMAP. 
Dense point clouds are produced by sequentially running following commands: (1) \textit{feature\_extractor}, (2) \textit{exhaustive\_matcher}, (3) \textit{patch\_match\_stereo}, and (4) \textit{stereo\_fusion}. Given dense point clouds, meshes are produced by (5) \textit{poisson\_mesher}.

{\bf UNISURF\cite{oechsle2021unisurf}}. The quantitative and qualitative results in the paper are provided by the authors of UNISURF. 

\section{Additional Experimental Results}

\begin{figure}[!t]
  \includegraphics[width=\linewidth]{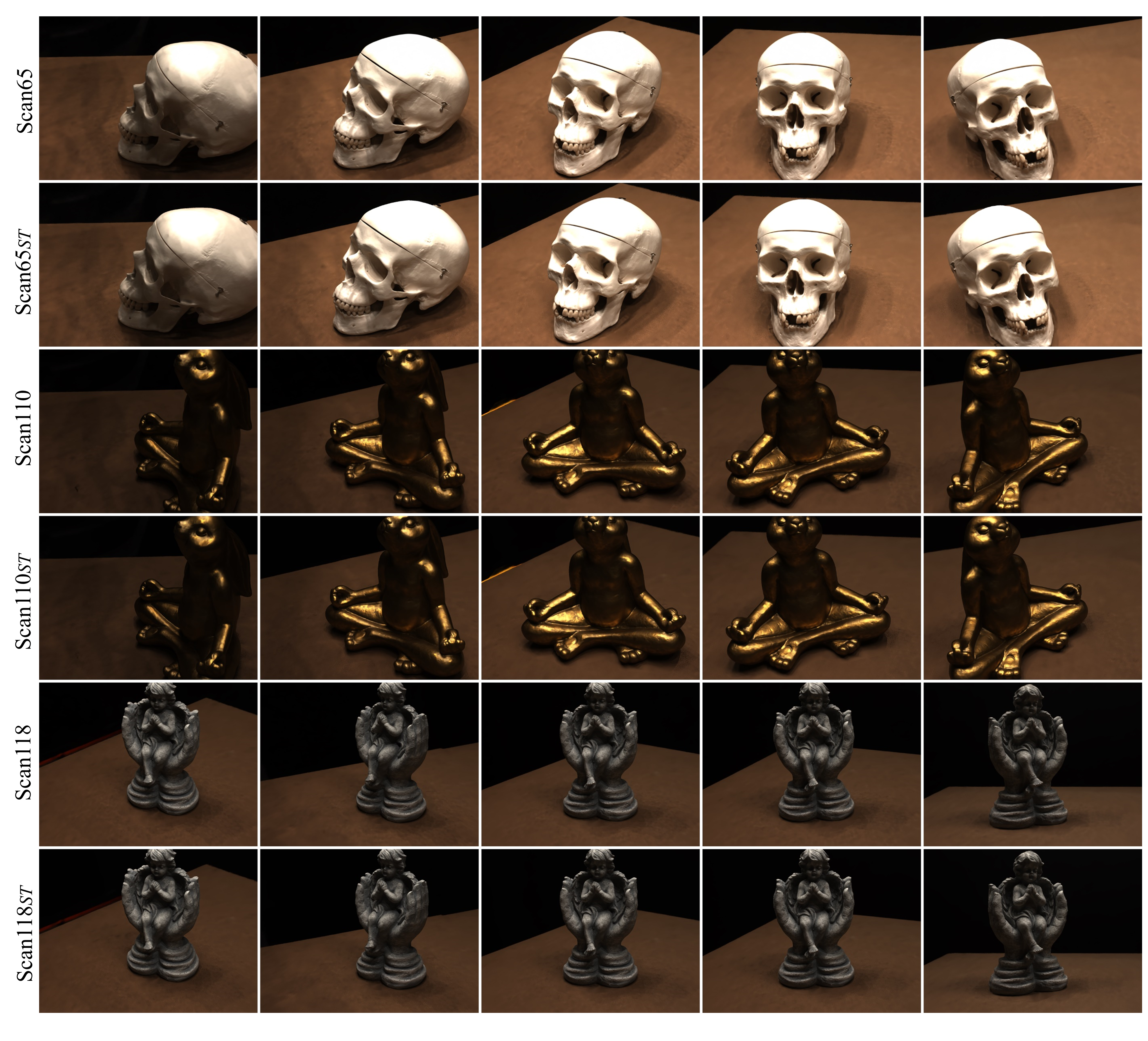}\\
  \caption{Rendered images by our method on the DTU dataset using different sampling strategies. {\small -$ST$} indicates the sampling strategy using sphere tracing.}
  \label{fig:rendering}
\end{figure}

\begin{table}[b]
    \hspace{-10pt}
    \setlength\tabcolsep{6pt} % default value: 6pt
    \small
    \begin{tabular}{c}
        \begin{adjustbox}{max width=1.0\textwidth}
        \aboverulesep=0ex
        \belowrulesep=0ex
        \renewcommand{\arraystretch}{1.5}
        \setlength{\tabcolsep}{0.3em}
        \begin{tabular}[t]{l||ccccccccccccccc|c}
            \textbf{Scan ID} &
            24 & 37 & 40 & 55 & 63 & 65 & 69 & 83 & 97 & 105 & 106 & 110 & 114 & 118 & 122 & Mean\\
            \hline
            PSNR\small{(Ours)} & 28.20 & 27.10 & 28.13 & 28.80 & 32.05 & 33.75 & 30.96 & 34.47 & 29.57 & 32.98 & 35.07 & 32.74 & 31.69 & 36.97 & 37.07 & 31.97 \\
            PSNR\small{(Ours$_{ST}$)} & 27.07 & 26.58 & 27.70 & 28.37 & 31.32 & 31.39 & 30.20 & 31.79 & 28.58 & 30.87 & 33.61 & 32.40 & 31.33 & 35.55 & 35.96 & 30.85 \\
            \hline
            SSIM\small{(Ours)} & 0.764 & 0.813 & 0.737 & 0.768 & 0.917 & 0.835 & 0.845 & 0.850 & 0.837 & 0.837 & 0.875 & 0.876 & 0.861 & 0.891 & 0.892 & 0.840 \\
            SSIM\small{(Ours$_{ST}$)} & 0.757 & 0.811 & 0.736 & 0.759 & 0.915 & 0.788 & 0.813 & 0.812 & 0.794 & 0.811 & 0.852 & 0.862 & 0.847 & 0.867 & 0.873 & 0.820
            \end{tabular}
        \end{adjustbox}
        \vspace{3pt}
    \end{tabular}
    \caption{Quantitative comparisons by different sampling strategies. {\small -$ST$} indicates the sampling strategy with sphere tracing.} 
    \label{tbl:sphere_tracing}
\end{table}

\begin{table}[htb]
    \hspace{-10pt}
    \setlength\tabcolsep{6pt} % default value: 6pt
    \small
    \begin{tabular}{c}
        \begin{adjustbox}{max width=1.0\textwidth}
        \aboverulesep=0ex
        \belowrulesep=0ex
        \renewcommand{\arraystretch}{1.5}
        \setlength{\tabcolsep}{0.3em}
        \begin{tabular}[t]{l||ccccccccccccccc|c}
            \textbf{Scan ID} &
            24 & 37 & 40 & 55 & 63 & 65 & 69 & 83 & 97 & 105 & 106 & 110 & 114 & 118 & 122 & Mean\\
            \hline
            PSNR\small{(NeRF)} & 24.83 & 25.35 & 26.87 & 27.64 & 30.24 & 29.65 & 28.03 & 28.94 & 26.76 & 29.61 & 32.85 & 31.00 & 29.94 & 34.28 & 33.69 & 29.31 \\
            PSNR\small{(Ours)} & 23.98 & 22.79 & 25.21 & 26.03 & 28.32 & 29.80 & 27.45 & 28.89 & 26.03 & 28.93 & 32.47 & 30.78 & 29.37 & 34.23 & 33.95 & 28.55 \\
            \hline
            SSIM\small{(NeRF)} & 0.753 & 0.794 & 0.780 & 0.761 & 0.915 & 0.805 & 0.803 & 0.822 & 0.804 & 0.815 & 0.870 & 0.857 & 0.848 & 0.880 & 0.879 & 0.826 \\
            SSIM\small{(Ours)} & 0.732 & 0.778 & 0.722 & 0.739 & 0.915 & 0.809 & 0.818 & 0.831 & 0.812 & 0.815 & 0.866 & 0.863 & 0.847 & 0.878 & 0.878 & 0.820
            \end{tabular}
        \end{adjustbox}
        \vspace{3pt}
    \end{tabular}
    \caption{Quantitative comparisons with NeRF on the task of novel view synthesis without mask supervision. \vspace{-17pt}} 
    \label{tbl:nvs}
\end{table}

\subsection {Rendering Quality and Speed}
Besides the reconstructed surfaces, our method also renders high-quality images, as shown in \figref{rendering}. Rendering an image in resolution of 1600x1200 costs about 320 seconds in the default volume rendering setting on a single Nvidia 2080Ti GPU. In addition, we also tested another sampling strategy by first applying sphere tracing to find the regions near the surfaces and only sampling points in those regions. With this strategy, rendering an image in the same resolution only needs about 60 seconds. Table~\ref{tbl:sphere_tracing} reports the quantitative results in terms of PSNR and SSIM in default volume rendering setting and sphere tracing setting.

\subsection {Novel View Synthesis}
In this experiment, we held out 10\% of the images in the DTU dataset as the testing set and the others as the training set. We compare the quantitative results on the testing set in terms of PSNR and SSIM with NeRF. As shown in Table~\ref{tbl:nvs}, our method achieves comparable performance to NeRF.

\begin{figure}[!b]
  \includegraphics[width=\linewidth]{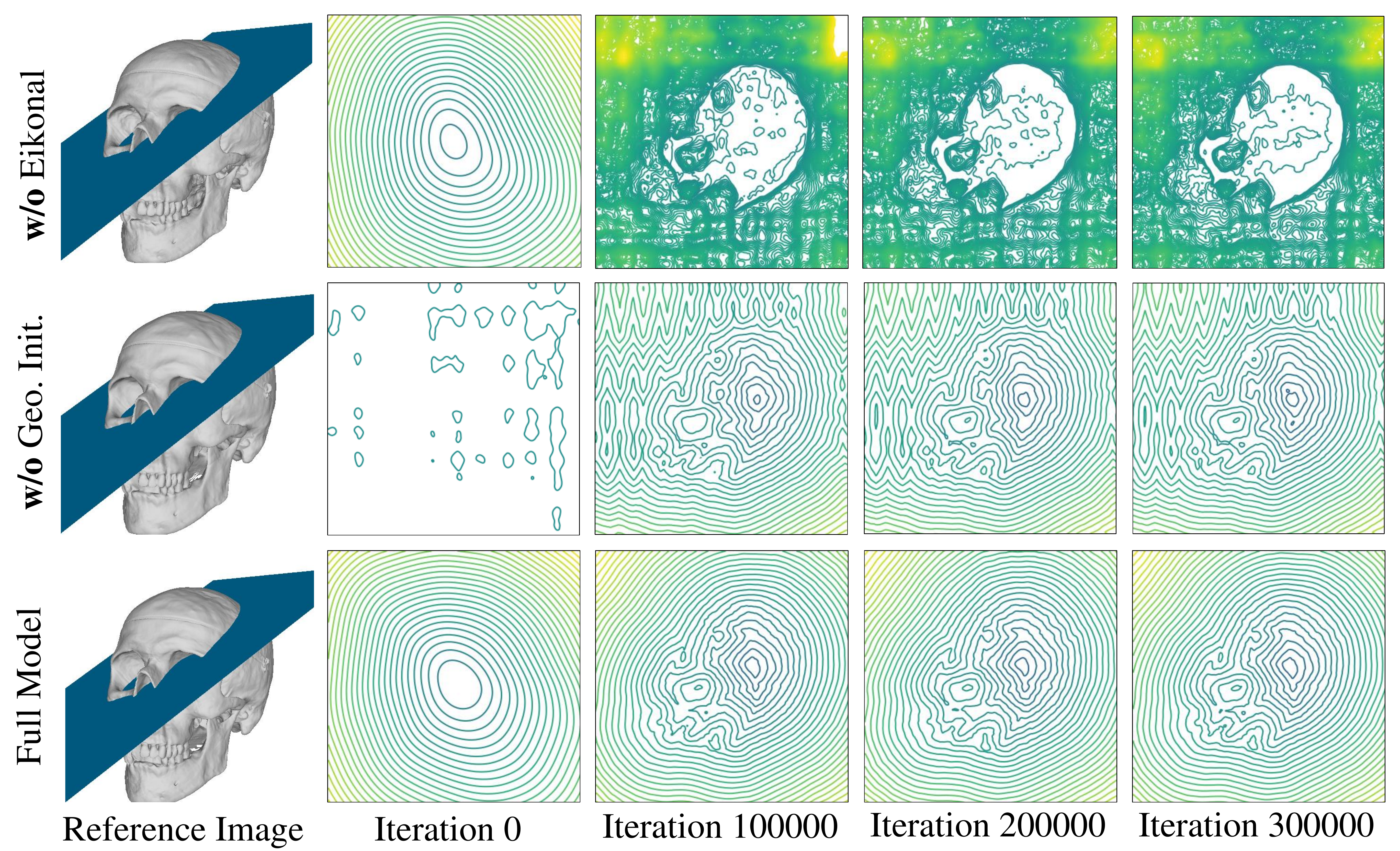}\\  \caption{Visualization of signed distance fields on the cutting plane (blue plane of the left image) in different training iterations.}
  \label{fig:sdf}
\end{figure}

\begin{figure}[!t]
  \includegraphics[width=\linewidth]{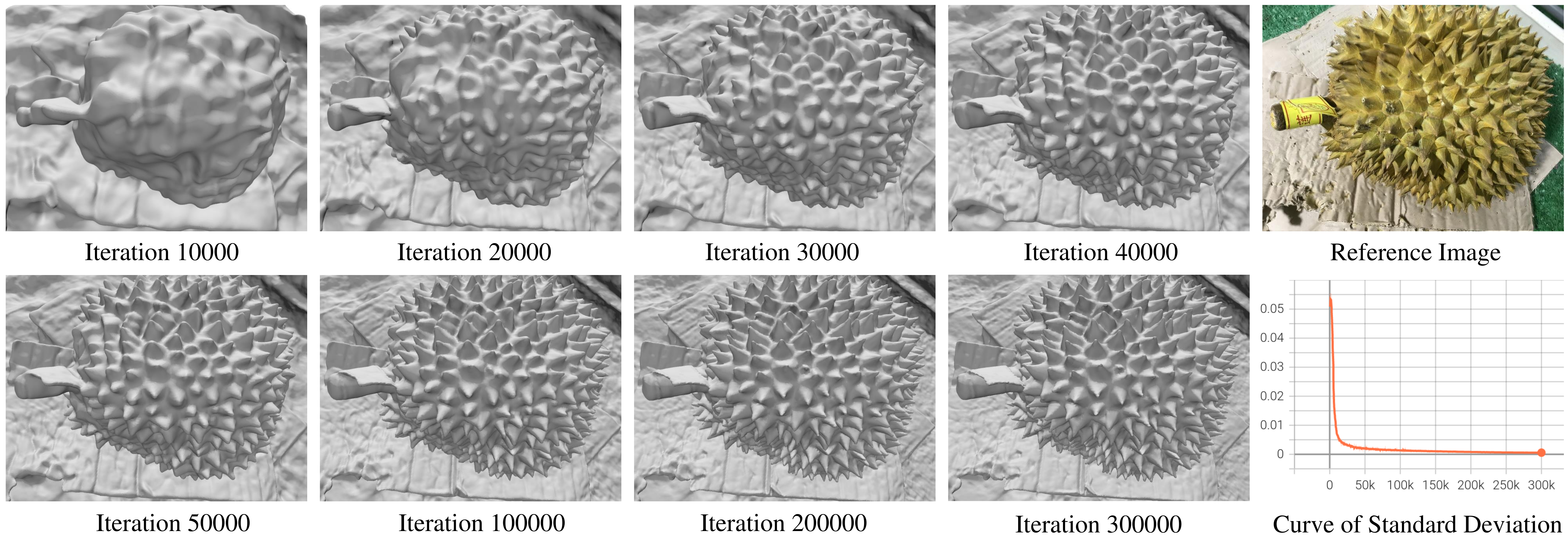}\\
  \caption{Training progression of the Durian in the BlendedMVS dataset. The bottom right figure shows the curve of the trainable standard deviation in the training progress.}
  \label{fig:iteration}
\end{figure}

\begin{figure}[b!]
  \includegraphics[width=\linewidth]{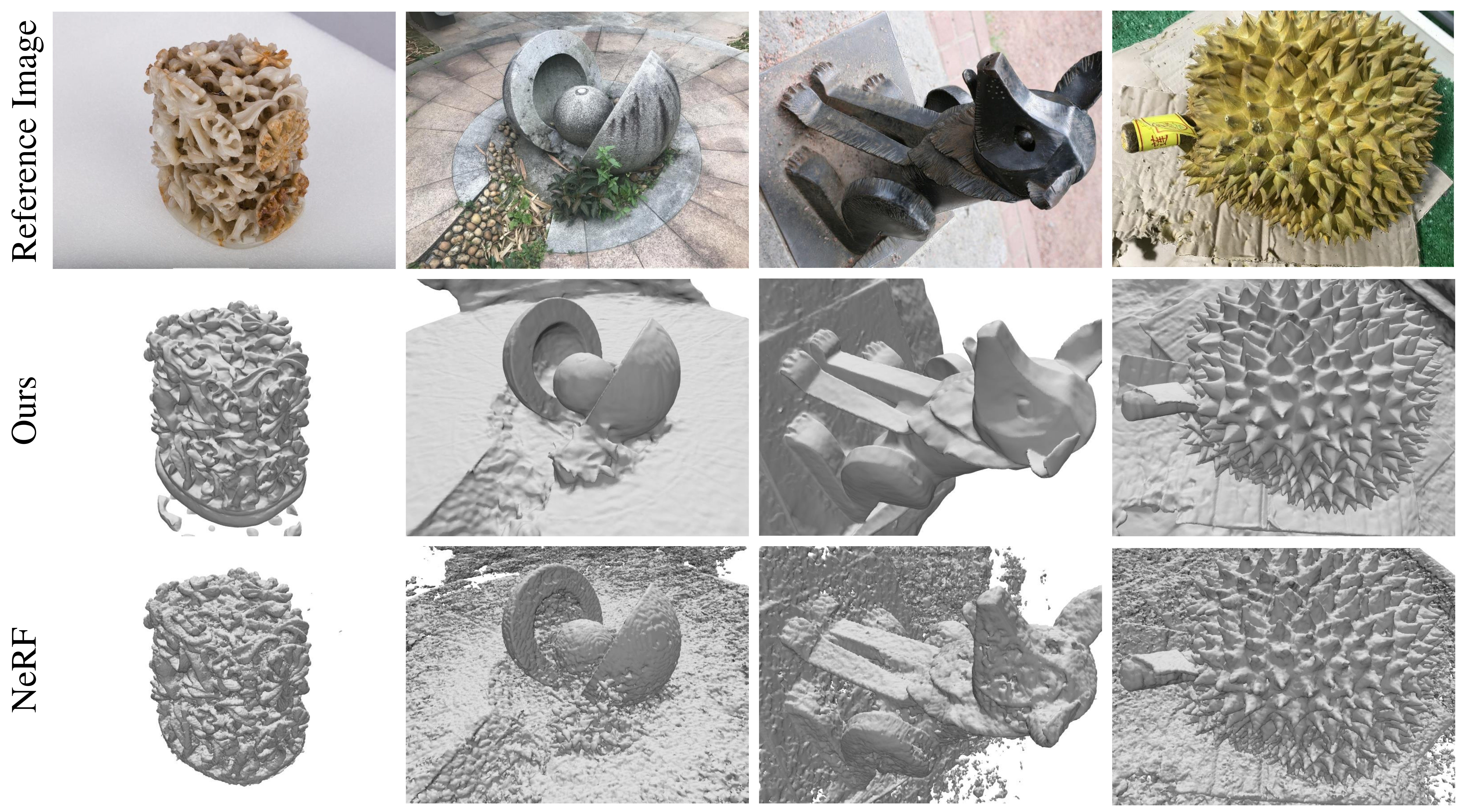}\\
  \caption{Additional reconstruction results on BlendedMVS dataset without mask supervision.}
  \label{fig:blendmvs}
  %\vspace{500pt}
\end{figure}

\subsection{SDF Qualitative Evaluation}
While our method without Eikonal regularization~\cite{gropp2020implicit} or geometric initialization~\cite{atzmon2020sal} produces plausible surface reconstruction results, our full model can predict a more accurate signed distance function as shown in \figref{sdf}. Furthermore, using random initialization produces axis-aligned artifacts due to the spectral bias of positional encoding~\cite{tancik2020fourier} while the geometric initialization~\cite{atzmon2020sal} does not have such kind of artifacts.

\subsection{Training Progression} We show the reconstructed surfaces at different training stages of the Durian in the BlendedMVS dataset. As illustrated in \figref{iteration}, the surface gets sharper along the training process. Meanwhile, we also provide a curve in the figure to show how the trainable standard deviation in $\phi_s$ changes in the training process. As we can see, the optimization process will automatically reduce the standard deviation so that the surface becomes more clear and sharper with more training steps.

\begin{wrapfigure}{R}{0.7\textwidth}
\centering{
  \vspace{-20pt}
  \includegraphics[width=\linewidth]{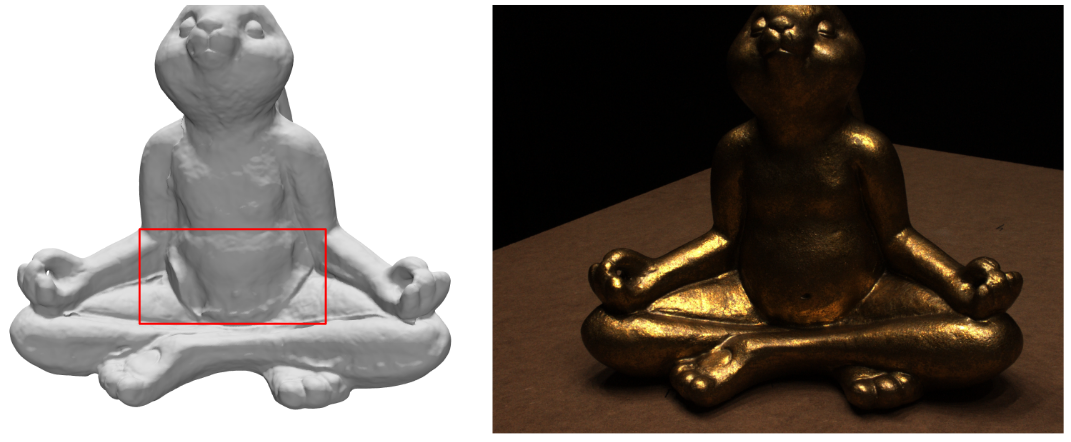}
  \caption{A failure reconstruction case containing textureless regions.}
  \vspace{-15pt}
  \label{fig:limitation}
}
\end{wrapfigure}

\subsection{Limitation}
\figref{limitation} shows a failure case where our method fails to correctly reconstruct the texutreless region of the surface on the metal rabbit model. The reason is that such textureless regions are ambiguous for reconstruction in neural rendering. 

\subsection{Additional Results}
\begin{figure}[htb]
  \includegraphics[width=\linewidth]{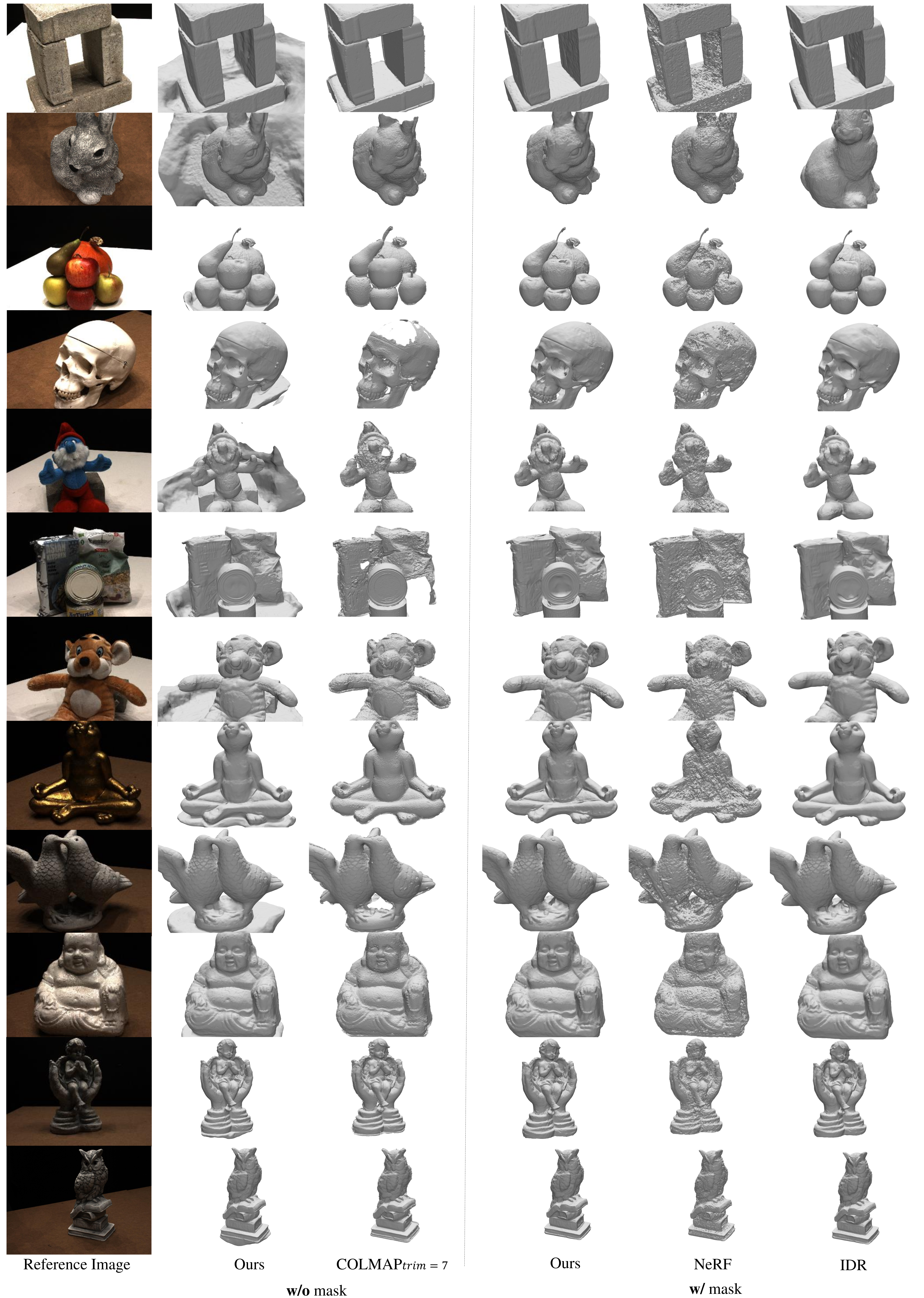}\\
  \caption{Additional reconstruction results on the DTU dataset.}
  \label{fig:dtu}
\end{figure}

In this section, we show additional qualitative results on the DTU dataset and BlendedMVS dataset. \figref{dtu} shows the comparisons with baseline methods in both \textbf{w/} mask setting and \textbf{w/o} mask setting. \figref{blendmvs} shows addtional results in \textbf{w/o} mask setting.

\end{document}